\documentclass[runningheads]{llncs}

\usepackage[T1]{fontenc}
\usepackage[utf8]{inputenc}
\usepackage{amsmath, amssymb}
\usepackage{graphicx}
\usepackage{tikz}
\usetikzlibrary{positioning, fit, backgrounds, calc, shapes.geometric}
\usepackage{booktabs}
\usepackage{microtype}
\usepackage[hidelinks]{hyperref}

\spnewtheorem{defn}{Definition}{\bfseries}{\itshape}

\begin{document}

\title{Analytic Abduction: Causal Decomposition and Governed Commitment for Human--AI Coordination\thanks{Accepted for presentation at the International Conference on Human and Artificial Rationalities (HAR 2026). The final authenticated version will be published in the conference proceedings in Springer's \emph{Lecture Notes in Computer Science} (LNCS) series.}}

\titlerunning{Analytic Abduction}

\author{Remo Pareschi\orcidID{0000-0002-4912-582X}}
\authorrunning{R. Pareschi}

\institute{Stake Lab, University of Molise, Italy \\
\email{remo.pareschi@unimol.it}}

\maketitle

\begin{abstract}
Abductive reasoning operates in two directions. The synthetic mode builds explanations from available hypotheses; the analytic mode, conversely, identifies the latent factors whose interaction accounts for a complex observed state. This paper develops the analytic mode as a non-greedy, risk-sensitive discipline of commitment, in which candidate factors coexist and interact, resolving into committed conclusions only when explicit governance conditions are met. The formal core is the $\kappa$--$\tau$ apparatus: $\kappa$ encodes the epistemic interaction among hypotheses, and $\tau$ sets a commitment threshold calibrated to the decision's stakes. The central contribution is the \emph{causal cluster} --- a structured object recording which latent factors participate in a decomposition, with what weights and interaction structure --- together with a two-level architecture (intra-cluster $\kappa^*$, inter-cluster $\kappa^{**}$) that guards against causal misattribution. Demonstrated in epidemiological crisis decomposition and adversarial cyber threat analysis, the framework's contribution to human--AI reasoning is the \emph{legibility of suspended decomposition} as a shared coordination object, providing structural resistance to premature convergence. In practice, the decision-maker is handed not a single imposed answer but the competing explanatory scenarios --- weighted by plausibility and paired with the evidence that would resolve between them --- so that sound action is possible even before the ambiguity is resolved.

\end{abstract}

\section{Introduction}

Abductive reasoning\footnote{The notion originates with C.\,S.\ Peirce, who distinguished abduction --- the generation of explanatory hypotheses --- from deduction and induction. We use \emph{abduction} throughout in this inference-to-explanation sense, without committing to the details of Peirce's own taxonomy.} operates in two directions. \emph{Synthetic abduction} composes hypotheses into explanations: given evidence, identify or construct the hypothesis whose interaction with the evidence best accounts for what has been observed. \emph{Analytic abduction} reverses this orientation: given a complex observed state of affairs, identify the latent factors whose interaction accounts for the observed complexity. The two modes share inferential substrate but differ in what they take as input, what they produce as output, and --- crucially --- in the risk profile of premature commitment.

This paper develops the analytic mode of the framework introduced in \cite{pareschi2025qa}, in which hypotheses --- drawn from the context as latent factors --- coexist, interact, and resolve into committed conclusions only when governance conditions are met. The analytic process is treated not merely as abstract inference but as group decision-making, in which the group may comprise both human and artificial agents, and in which the explanations reached and the decisions they license must satisfy previously agreed criteria of evidence and validity. Because explanatory elements are acquired concurrently, consolidated hypotheses interact dynamically rather than remaining statically isolated.

This mechanism of dynamic interaction --- articulated as semantic interference in the entangled-heuristics architecture of \cite{ghisellini2025entangled} for strategic reasoning --- was generalised by \cite{pareschi2025qa} into the Quantum Abduction (QA) framework,\footnote{The qualifier ``quantum'' is used in the sense of \emph{quantum cognition}: mathematical structures borrowed from quantum theory (superposition-like coexistence of alternatives, interference between them) are used to model graded, interacting plausibility. It carries no claim that cognition is physically quantum-mechanical. The present paper develops the analytic mode through the $\kappa$--$\tau$ governance mechanics and does not depend on this metaphor.} formalising the $\kappa$-interaction and $\tau$-commitment dynamics across forensic, clinical, and strategic domains; there, quantum superposition served as a conceptual scaffold for hypothesis coexistence and synthesis. The present paper retains these governance mechanics while developing them in the dual, analytic direction --- decomposition rather than synthetic composition --- and demonstrating their operation in two domains where premature commitment carries severe downside cost: crisis decomposition under epidemiological uncertainty, and adversarial strategy decomposition in cyber threat analysis.

\subsection*{Two Orthogonal Distinctions}

We position the paper within two distinctions that structure the contribution.

The first is the distinction between \emph{abduction as process} and \emph{the $\kappa$--$\tau$ apparatus as logic}. Abduction in complex settings --- where hypotheses co-occur, are acquired concurrently, and interact rather than being evaluated in isolation, the sense in which \cite{pareschi2025qa} termed the process \emph{quantum} --- names, as developed there and extended here, the multi-agent socio-technical activity of reasoning under uncertainty across human decision-makers, large language models, and specialised AI components. The $\kappa$--$\tau$ apparatus --- the pair of governance parameters comprising the epistemic interaction operator $\kappa$ and the normative commitment threshold $\tau$ --- is the formal discipline that structures this process. The distinction matters because it locates the present paper's contribution precisely: this is a paper about the analytic mode of QA in its agentic form, with the $\kappa$--$\tau$ apparatus serving as the inferential substrate that the agentic process inhabits.

The second is the distinction between \emph{synthetic} and \emph{analytic} abduction introduced above. The synthetic mode composes upward from a library of atomic hypotheses toward emergent composite explanations. The analytic mode operates in the dual direction: given a structured explanandum $\Phi$ and a domain-specific library of latent factors $\mathcal{F}$, the framework decomposes $\Phi$ into candidate \emph{causal clusters} --- structured objects recording which latent factors participate in a decomposition, with what weights, and in what interaction structure. The central formal contribution of the paper is the apparatus of causal clusters together with a two-level interaction architecture: intra-cluster interaction within a candidate decomposition, inter-cluster interaction between competing decompositions, and $\tau$ governing commitment at either level. This contribution is conditioned on its inputs: the factor library $\mathcal{F}$ and the LLM-based candidate-generation stage bound what any decomposition can contain --- a dependence we make explicit in \S6 and return to in the Limitations.

\subsection*{The Distinctive Contribution to Human-AI Reasoning}

The paper's contribution to the structure of human-AI reasoning is not located in any single locus of the analytic-abduction pipeline. The LLM proposes candidate decompositions; domain experts calibrate interaction parameters; institutional decision-makers set the commitment threshold $\tau$ to encode the risk profile of the domain; specialized AI components contribute pattern recognition and projection at scale. These distributed contributions are individually unremarkable: any architecture for human-AI reasoning under uncertainty distributes labor across human and AI components in similar ways.

What is distinctive is what the $\kappa$--$\tau$ apparatus makes representable across this distributed pipeline: \emph{the legibility of suspended decomposition as a shared coordination object}. When the framework determines that multiple candidate decompositions are plausible but none is commit-worthy --- that competing causal clusters cannot yet be disambiguated by the available evidence --- the output is not a probabilistic distribution to be interpreted, nor a confidence-ranked list to be acted upon, nor a refusal to answer. The output is a \emph{structured account of the current inferential state}: which decompositions are plausible, how they compete, why no commitment is warranted, and what evidence would resolve the ambiguity. This structured non-commitment is itself a coordination object that human and AI agents at different points in the multi-agent system can read, contribute to, and act on. Without such a shared object, multi-agent systems are vulnerable to \emph{premature convergence}: whichever agent first reaches a confident conclusion drives collective commitment toward that conclusion regardless of whether the inferential state warrants it. The $\kappa$--$\tau$ apparatus provides structural resistance to this failure mode by making non-commitment as legible as commitment.

\subsection*{Adversarial Reasoning as a Sharpening Case}

The two demonstrations of the paper exhibit a deliberate progression. The first --- crisis decomposition in an epidemiological setting (\S4) --- exhibits all the structural features that motivate the analytic mode: complex explanandum, multiple plausible decompositions, severe misdecomposition cost, and the value of legible suspended decomposition for action guidance. In this setting, the pathogen does not anticipate the analyst; the framework operates on a generatively innocent phenomenon.

The second demonstration (\S5) introduces what the first does not: an adversary that anticipates the analyst's decomposition and may craft observable behavior to mislead it. Under adversarial conditions, premature commitment is no longer merely costly --- it becomes an \emph{attack surface}, exploitable by an adversary sophisticated enough to model the analyst's causal attribution. The cyber threat-analysis setting is the domain in which this sharpening becomes operationally severe. It also surfaces a phenomenon we call \emph{structural-versus-linguistic novelty}: an adversary may use entirely familiar tools (high projection on known factors) while combining them in compositionally novel ways (anomalous intra-cluster interaction pattern). The $\kappa$-theoretic framing supplies the formalization of this distinction, which conventional similarity-based attribution conflates.

\subsection*{Plan of the Paper}

\S2 recapitulates the elements of \cite{ghisellini2025entangled,pareschi2025qa} load-bearing for the analytic extension and introduces the $\kappa$--$\tau$ apparatus as a named pair of governance parameters. \S3 develops the formal core of the analytic mode: the causal cluster as the structured output of decomposition, the cluster-score semantics, the two-level interaction architecture, and the $\tau$-governed separation condition that distinguishes commit-worthy decompositions from suspended ones. \S4 demonstrates the apparatus in epidemiological crisis decomposition. \S5 develops the adversarial demonstration in cyber threat analysis, including the structural-versus-linguistic novelty phenomenon. \S6 generalizes the human-AI coordination argument: the legibility of suspended decomposition as a coordination object and a structural defense against premature convergence. \S7 situates the contribution within related work. \S8 concludes with limitations and forward connections to companion work.

\section{Background: The $\kappa$--$\tau$ Apparatus}

The framework builds on the entangled-heuristics architecture of \cite{ghisellini2025entangled}, generalised in \cite{pareschi2025qa} into the Quantum Abduction (QA) framework. This section gives only the technical background the analytic extension requires: the semantic substrate on which projection is computed (\S2.1), the strategic-heuristics architecture that supplies the factor library and coordination pattern used later (\S2.2), and the two governance parameters $\kappa$ and $\tau$ (\S2.3). The broader orientation --- abduction as a multi-agent, socio-technical process rather than a head-internal inference --- was set out in the Introduction and is not repeated here.

\subsection{The Semantic Substrate}

Each hypothesis $h_i$ --- in analytic mode, each factor $f_i$ --- and each observation $o_j$ is represented as a vector in a shared semantic space, and the activation of $h_i$ by $o_j$ is a projection score derived from $\cos(\mathbf{h}_i, \mathbf{o}_j)$. Hypotheses do not act in isolation: an interaction structure (\S2.3) lets co-active hypotheses reinforce or attenuate one another, and a synthesis operator composes them into composite explanations when their interaction warrants it --- the operator made precise, in the analytic direction, in \S3.3. Commitment occurs only when a score exceeds the threshold $\tau$ (\S2.3); below it, hypotheses remain active and open to revision. This is a vector-space realisation of graded, interacting plausibility, and presupposes no quantum-mechanical claim about cognition.

\textbf{The embedding layer as modular substrate.} The realisation in \cite{ghisellini2025entangled,pareschi2025qa} uses Sentence-BERT \cite{reimers2019sbert}, but the framework treats the embedding layer as a substitutable component: any model mapping text to vectors in a shared space supplies the cosine projection the $\kappa$--$\tau$ apparatus consumes, and the layer can be swapped --- for a contradiction-aware variant such as SimCSE \cite{gao2021simcse}, or a domain-adapted encoder --- without altering the inferential pipeline. Better embeddings yield more discriminating projections and $\kappa$ values, with no change to the apparatus. The worked examples of \S\S4--5 exploit exactly this modularity.

\subsection{Entangled Heuristics for Strategic Reasoning}

The entangled-heuristics architecture of \cite{ghisellini2025entangled} is the concrete realisation on which QA was originally based. Its elements are not hypotheses about a determinate reality but \emph{heuristics} --- conditional principles of action drawn from strategic traditions --- embedded as vectors, with interaction computed through a matrix of coefficients $\kappa_{ij}$ and composition delegated to a large language model under structured prompting; conflicting heuristics are fused rather than eliminated. Two features carry over directly: the heuristics library, extended with operationally-grounded offensive-security heuristics \cite{carapella}, supplies the factor library $\mathcal{F}$ of \S5; and the architectural pattern --- LLM-mediated composition over a semantic substrate with inspectable interaction matrices --- shapes the human--AI coordination patterns of \S6.

\subsection{The $\kappa$--$\tau$ Apparatus}

Two quantities govern the dynamics of explanatory states. The \emph{interaction parameter} $\kappa(h_i, h_j) \in [-1, 1]$ encodes the qualitative character of the interaction between two hypotheses (or, in analytic mode, two factors): positive values indicate reinforcement, negative values attenuation, and values near zero independence. It is estimated from the semantic similarity of the embedded representations, refined by domain experts, and where available validated against cases.

The \emph{commitment threshold} $\tau$ separates having a plausible conclusion from being committed to it. A hypothesis or composite whose score exceeds $\tau$ is \emph{commit-worthy} --- eligible for inferential closure and for licensing downstream action; one whose score is positive but below $\tau$ is \emph{plausible but suspended} --- epistemically active and open to further interaction, but not yet a warranted basis for commitment. The threshold is calibrated to the stakes of the domain: high-stakes settings, where acting on a wrong conclusion costs far more than continued investigation, warrant a high $\tau$ and defer commitment until evidence is strong.

The significance of the pair lies in what it makes representable. $\kappa$ introduces compositional interaction among hypotheses, and $\tau$ structurally decouples derivation from commitment --- so that inference may legitimately continue past the point where a probabilistically-coherent system would have closed, because the information generated by sustained interaction among active hypotheses is itself inferentially valuable. \S3 develops the analytic mode within this apparatus.

\section{From Synthetic to Analytic Abduction}

The synthetic mode, as developed in \cite{pareschi2025qa}, reasons \emph{upward}: a finite library of atomic hypotheses $H = \{h_1, \ldots, h_n\}$ is composed, under interaction, into emergent explanations whose plausibility exceeds that of any single component. Given observations $O$, the framework activates hypotheses through projection, holds several jointly active prior to commitment, and --- when their interaction is sufficiently reinforcing --- synthesizes hybrids that no single hypothesis could produce in isolation. This is the \emph{synthetic} mode: hypotheses are the starting material, the composite explanation is the output.

This section develops the dual orientation. In analytic abduction, the starting material is a complex observed state, and the output is a structured decomposition: a set of latent factors whose interaction accounts for the observed complexity. Where synthetic abduction asks \emph{which combination of known hypotheses best explains the evidence}, analytic abduction asks \emph{what hidden decomposition --- which interacting latent factors, with what compatibility structure --- best accounts for the observed state}. The formal substrate transfers: the $\kappa$--$\tau$ apparatus that governs synthesis in QA also governs decomposition in analytic mode. What changes is the interpretive orientation, the structure of the output, and --- critically --- the risk profile of premature commitment.

\subsection{Inputs to Analytic Abduction}

The synthetic mode of QA, as developed in \cite{pareschi2025qa}, operates on two inputs: a set of observations $O = \{o_1, \ldots, o_m\}$ and a hypothesis library $H = \{h_1, \ldots, h_n\}$ from which composite explanations are constructed. The analytic mode reorients both inputs.

In place of discrete observations, analytic abduction operates on a structured \emph{explanandum}: a single observed phenomenon presented as an integrated complex state, denoted $\Phi$. The explanandum is the analytic counterpart of an integrated pattern: an outbreak trajectory rather than a set of individual case reports; an adversary campaign rather than a set of isolated alerts; a systemic correlation pattern rather than a set of independent indicator readings. The framework takes $\Phi$ as primitive; how it was constituted from upstream data --- whether by domain experts, by pattern-recognition systems, or by the multi-agent collective in which the framework is embedded --- is outside the scope of the decomposition layer and is treated, in \S6, as part of the surrounding agentic process.

In place of a hypothesis library, analytic abduction operates on a \emph{factor library} $\mathcal{F} = \{f_1, f_2, \ldots\}$: a domain-specific catalogue of latent causal elements that the system is willing to consider as candidates for decomposition. Epidemiology supplies transmission vectors; strategic theory supplies the heuristics catalogued in \cite{ghisellini2025entangled}; financial risk analysis supplies the channels through which contagion propagates. The factor library is structurally analogous to the hypothesis library of synthetic mode, but its elements play a different role: factors are not themselves explanations of $\Phi$, but the \emph{components} whose interaction, captured in a causal cluster, constitutes a candidate decomposition.

The dual orientation is now precise. Synthetic abduction takes $\langle O, H \rangle$ as input and produces a composite hypothesis whose plausibility is scored against $O$. Analytic abduction takes $\langle \Phi, \mathcal{F} \rangle$ as input and produces a causal cluster $C = \langle w, \kappa_C \rangle$ over $\mathcal{F}$ whose cluster score $\mathrm{sc}_S(C \mid \Phi)$ measures how well the synthesized factors account for the explanandum.

The framework presupposes that $\mathcal{F}$ contains the operative factors, or compositions thereof. Decomposition under incomplete factor libraries --- where the operative factor is genuinely novel and not synthesizable from existing $\mathcal{F}$ --- falls outside the present framework and is flagged as future work in \S8.

\subsection{The Explanandum as a Structured Object}

The cluster score $\mathrm{sc}_S(C \mid \Phi)$ introduced in \S3.3 below presupposes that factor projections against $\Phi$ are well-defined. In the synthetic mode of QA \cite{pareschi2025qa}, observations are individual textual or propositional items embedded as vectors in a semantic space, and the projection $\alpha_i = \cos(\mathbf{h}_i, \mathbf{o}_j)$ is the standard cosine between two vectors. In analytic mode, the explanandum is not a single observation but an \emph{integrated complex state}. Treating $\Phi$ as a single mean vector would destroy the structural distinction that motivates the analytic mode in the first place --- the whole point of decomposition is that the explanandum exhibits \emph{structure} beyond what its constituent observations individually carry.

We therefore encode $\Phi$ as a structured tuple
\[
\Phi = \langle O_\Phi,\; R_\Phi,\; \mu_\Phi \rangle
\]
with three components:

\begin{itemize}
  \item $O_\Phi = \{o_1, \ldots, o_m\}$ is the set of \emph{constituent observations}: the individual items whose joint pattern constitutes the explanandum. Each $o_j$ is itself embeddable in the semantic space following \cite{pareschi2025qa}, yielding a vector $\boldsymbol{o}_j \in \mathbb{R}^d$.
  \item $R_\Phi$ is the \emph{relational structure} over $O_\Phi$: a representation of how the observations relate to one another. Relations may include temporal ordering, spatial co-occurrence, causal precedence as inferred by upstream systems, or domain-specific relations such as actor-target pairing in adversarial settings or transmission-chain membership in epidemiological settings. Formally, $R_\Phi$ is a labeled relation $R_\Phi \subseteq O_\Phi \times O_\Phi \times L$ for some label set $L$ encoding relation types; alternatively a hypergraph for higher-arity relations.
  \item $\mu_\Phi : O_\Phi \to [0, 1]$ is a \emph{salience measure} assigning each observation a weight reflecting its evidential prominence within the explanandum: how heavily $o_j$ figures in the pattern that constitutes $\Phi$.
\end{itemize}
\textbf{Projection of factors against the explanandum.} Given a factor $f_i \in \mathcal{F}$ with embedding $\mathbf{f}_i$ and an explanandum $\Phi = \langle O_\Phi, R_\Phi, \mu_\Phi \rangle$, the projection of $f_i$ against $\Phi$ is defined through a \emph{relational aggregator} $\Psi$:
\[
\alpha_i(\Phi) \;:=\; \Psi\!\left( \{ \cos(\mathbf{f}_i, \boldsymbol{o}_j) \cdot \mu_\Phi(o_j) \}_{o_j \in O_\Phi},\; R_\Phi \right)
\]
The aggregator $\Psi$ takes two inputs: the individual cosine projections of $f_i$ against each constituent observation, weighted by salience; and the relational structure $R_\Phi$. It returns a scalar in $[0, 1]$ representing the projection of $f_i$ against $\Phi$ as a structured whole.

The framework does not prescribe the form of $\Psi$. Its simplest realization is the salience-weighted mean
\[
\Psi_{\min}(X, R) \;=\; \frac{1}{|X|} \sum_{j} x_j
\]
which ignores $R_\Phi$ and recovers a treatment of $\Phi$ as an unstructured set of weighted observations. Richer realizations exploit $R_\Phi$ structurally --- for example, by computing projections over relational substructures (temporal windows, co-occurrence cliques, causal chains) and aggregating with structure-sensitive weighting; or by employing graph-kernel-based aggregators that respect $R_\Phi$'s relational topology. As with $\kappa$ in the synthetic mode of \cite{pareschi2025qa}, the framework presupposes that $\Psi$ is supplied by domain instantiation; specific realizations are developed in \S\S4 and 5.

\textbf{Why the structured encoding matters.} Two consequences of the encoding warrant explicit notice.

First, the structural-versus-linguistic novelty phenomenon introduced in \S5.3 acquires a precise formal anchor. Linguistic novelty is novelty in $O_\Phi$ --- observations whose embeddings $\boldsymbol{o}_j$ project weakly onto the factor library. Structural novelty is novelty in $R_\Phi$ --- relations among observations that do not match the relational patterns of any previously catalogued case. The two are mathematically independent: a campaign can exhibit familiar individual observations (high cosine $\cos(\mathbf{f}_i, \boldsymbol{o}_j)$ for many $i, j$) combined under an unfamiliar relational structure $R_\Phi$, and the framework's aggregator $\Psi$ can in principle detect this dissociation while a treatment of $\Phi$ as a mean vector cannot.

Second, the upstream constitution of $\Phi$ discussed in \S3.1 and developed in \S6.3 acquires formal content. The agents that contribute to explanandum formation --- domain experts, pattern-recognition systems, the multi-agent collective --- are not merely naming $\Phi$ but producing its components: identifying the constituent observations $O_\Phi$, inferring the relational structure $R_\Phi$, and calibrating the salience measure $\mu_\Phi$. Each component admits human, AI, or collaborative contribution. The legibility argument of \S6 then has a concrete target: agents downstream of explanandum formation can inspect $R_\Phi$ and $\mu_\Phi$ as well as $O_\Phi$, and can contribute corrections or refinements to any of the three.

The remainder of the formal development proceeds as before. Cluster scoring (\S3.3) and the two-level interaction architecture (\S3.5) operate over the structured explanandum: wherever cluster scoring and projection reference $\Phi$, the projection is $\alpha_i(\Phi)$ as defined above, and the synthesis operator $\otimes$ acts on factor projections defined through $\Psi$.

\subsection{The Causal Cluster}

In synthetic mode, the scoring semantics ultimately produces a scalar: the plausibility of a composite hypothesis $t_1 \otimes t_2$. This scalar suffices when the inferential question is \emph{how plausible is this composite explanation}. In analytic mode, the inferential question is structurally different: \emph{what is the decomposition}. A scalar plausibility score cannot answer this question; the answer requires preserving the structure that the score reduces.

We therefore introduce the \emph{causal cluster} as the structured output object of analytic abduction.

\begin{defn}[Causal cluster]
\label{def:causal-cluster}
Let $F = \{f_1, \ldots, f_k\} \subseteq \mathcal{F}$ be a set of latent factors drawn from the factor library. A \emph{causal cluster} over $F$ is a pair
\[
C = \langle w, \kappa_C \rangle
\]
where:
\begin{itemize}
  \item $w : F \to [0, 1]$ assigns each factor a weight reflecting its explanatory contribution within the decomposition;
  \item $\kappa_C : F \times F \to [-1, 1]$ specifies the internal interaction structure, encoding whether factors reinforce ($\kappa_C(f_i, f_j) > 0$), inhibit ($\kappa_C(f_i, f_j) < 0$), or operate independently ($\kappa_C(f_i, f_j) \approx 0$) within the cluster.
\end{itemize}
\end{defn}

The cluster $C$ thus records not merely \emph{which} factors are operative, but \emph{how} they combine. Two clusters may identify the same factors with the same weights yet differ in their internal $\kappa_C$ --- one representing a decomposition in which factors operate independently, another representing a decomposition in which the factors interact strongly. These are epistemically distinct decompositions with potentially divergent intervention implications, and the causal cluster preserves this distinction that scalar reduction would destroy.

\textbf{Scoring a cluster against an explanandum.} We do not introduce a new operator for decomposition. Instead, we reuse the compositional value semantics of \cite{pareschi2025qa} and reverse the inferential direction. The synthetic mode of \cite{pareschi2025qa} scores a composite by a value decomposition (its Eq.~2) that sums an \emph{individual} contribution of each constituent and a \emph{pairwise interaction} contribution modulated by the coupling between constituents. We read this decomposition in the analytic direction. Given a cluster $C = \langle w, \kappa_C \rangle$ over factors $F = \{f_1, \ldots, f_k\}$ and an explanandum $\Phi$, write $\alpha_i := \alpha_i(\Phi)$ for the projection of $f_i$ against $\Phi$ (\S3.2). Define the \emph{base fit} and the \emph{internal coherence} of the cluster,
\[
\mathrm{base}(C \mid \Phi) \;=\; \sum_{i=1}^{k} w_i\,\alpha_i,
\qquad
\mathrm{coh}(C) \;=\; \frac{\sum_{i<j}\, \kappa_C(f_i,f_j)\, w_i w_j\, \alpha_i \alpha_j}{\sum_{i<j}\, w_i w_j\, \alpha_i \alpha_j},
\]
where the weights are normalised within the cluster ($\sum_i w_i = 1$). The base fit is the salience-weighted projection of the cluster's factors onto $\Phi$ --- the individual-contribution term --- and lies in $[0,1]$. The coherence is the $\kappa_C$-weighted mean of the pairwise interactions actually exercised by the decomposition (each pair weighted by its joint explanatory mass $w_i w_j \alpha_i \alpha_j$), and lies in $[-1,1]$: positive when the cluster's reinforcing factor pairs carry the explanatory load, negative when its inhibitory pairs do. The \emph{cluster score} composes the two as a bounded modulation of fit by coherence,
\begin{equation}
\mathrm{sc}_S(C \mid \Phi) \;=\; \mathrm{base}(C \mid \Phi)\cdot \frac{1 + \eta\,\mathrm{coh}(C)}{1 + \eta},
\label{eq:cluster-score}
\end{equation}
with $\eta \in [0,1]$ a stated coupling constant governing how strongly internal coherence is allowed to raise or lower the bare fit. Equation~\eqref{eq:cluster-score} is the operator written $\bigotimes_{i} f_i$ in the schematic presentation above: it is defined for any arity $k$, reduces to the salience-weighted projection $\mathrm{base}(C\mid\Phi)$ when the cluster is interaction-free ($\kappa_C \equiv 0$), takes values in $[0, \mathrm{base}(C\mid\Phi)] \subseteq [0,1]$ without ad hoc clipping, and --- crucially for the role of the causal cluster --- separates two clusters that carry identical factors and weights but differ in $\kappa_C$, since they then differ in $\mathrm{coh}(C)$ and hence in score. The asymmetry between synthesis and decomposition is thus interpretive, not formal: both modes evaluate the same value decomposition of \cite{pareschi2025qa}; synthesis reads it as the plausibility of a composite, analytic abduction reads it as the adequacy of a decomposition.

This formulation preserves the duality cleanly: synthesis composes upward, analytic decomposition reverses the arrow but uses the same compositional machinery. The asymmetry between the two modes is interpretive, not formal.

\subsection{Multiple Clusters and Suspended Decomposition}

In synthetic mode, suspended derivation captures the epistemic state in which multiple composite hypotheses remain plausible without any single composite having crossed the commitment threshold. The analytic-mode counterpart is \emph{suspended decomposition}: multiple candidate clusters $\{C_1, C_2, \ldots, C_m\}$ remain plausible as decompositions of $\Phi$, and none has yet been committed to.

Formally, we say a cluster $C$ is \emph{plausible} relative to $\Phi$ if $\mathrm{sc}_S(C \mid \Phi) \geq \epsilon$, where $\epsilon$ is the activation floor inherited from \cite{pareschi2025qa}; we write
\[
\vdash^p_S C \mid \Phi.
\]
We say $C$ is \emph{commit-worthy} if $\mathrm{sc}_S(C \mid \Phi) \geq \tau$ and, additionally, no competing plausible cluster $C'$ exhibits sufficient explanatory rivalry to block commitment --- a condition we make precise in \S3.5 below. We write
\[
\vdash^c_S C \mid \Phi
\]
for the commit-worthy case.

The structural property that motivated the analogous distinction in synthetic mode now does double duty in analytic mode. A plausible cluster is not yet a committed decomposition; multiple plausible decompositions may coexist; and commitment to any one of them is a normative act regulated by $\tau$, not a logical consequence of having found a candidate that fits the explanandum.

\subsection{Two Levels of Interaction}

\begin{figure}[t]
\centering
\begin{tikzpicture}[
  factor/.style={circle, draw, minimum size=8mm, inner sep=0pt, font=\small},
  cluster/.style={rectangle, draw, dashed, rounded corners, inner sep=4mm, label={[font=\small\itshape]above:#1}},
  intra/.style={<->, thick, gray!70!black},
  inter/.style={<->, very thick, red!60!black, dashed},
  thresh/.style={font=\small\itshape, gray!60!black},
]
\node[factor] (f1) at (0, 0) {$f_1$};
\node[factor] (f2) at (1.4, 0.5) {$f_2$};
\node[factor] (f3) at (1.4, -0.5) {$f_3$};
\draw[intra] (f1) -- node[above, font=\tiny] {$\kappa^*_C$} (f2);
\draw[intra] (f1) -- (f3);
\draw[intra] (f2) -- (f3);
\node[cluster=Cluster $C_1$, fit=(f1) (f2) (f3)] (C1) {};

\node[factor] (g1) at (5.5, 0) {$f_4$};
\node[factor] (g2) at (6.9, 0.5) {$f_2$};
\node[factor] (g3) at (6.9, -0.5) {$f_5$};
\draw[intra] (g1) -- (g2);
\draw[intra] (g1) -- (g3);
\draw[intra] (g2) -- (g3);
\node[cluster=Cluster $C_2$, fit=(g1) (g2) (g3)] (C2) {};

\draw[inter] (C1) -- node[above, font=\small] {$\kappa^{**}(C_1, C_2)$} (C2);

\node[thresh] at (3.45, -1.8) {$\tau$ governs commitment at either level; $\delta(\tau)$ enforces separation};
\end{tikzpicture}
\caption{The two-level interaction architecture of analytic abduction. Solid edges represent intra-cluster interaction $\kappa^*_C$ (the dynamics within a candidate decomposition); the dashed red edge represents inter-cluster interaction $\kappa^{**}(C_1, C_2)$ (the compatibility or competition between candidate decompositions). The threshold $\tau$ regulates commitment at both levels, while the separation margin $\delta(\tau)$ blocks commitment to $C_1$ when a structurally incompatible competitor $C_2$ has comparable cluster score.}
\label{fig:two-level-architecture}
\end{figure}
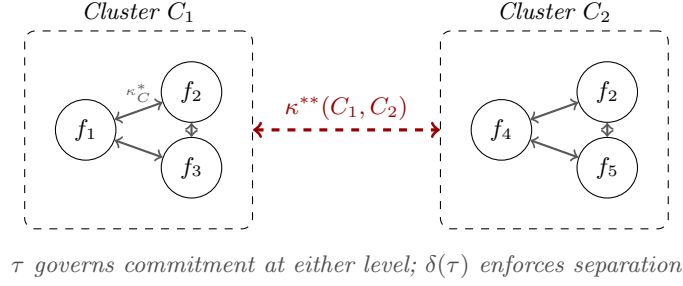

When multiple candidate clusters coexist under suspended decomposition, the interaction structure operates at two qualitatively different levels. Figure~\ref{fig:two-level-architecture} illustrates the architecture schematically.

\textbf{Intra-cluster interaction} ($\kappa^*$). Within a single cluster $C$, the factors interact via $\kappa_C$. The lifted interaction $\kappa^*$ inherited from \cite{pareschi2025qa} applies directly: the interaction between two factor terms within $C$ is the aggregate of the atomic $\kappa_C$ values across their constituents. This level governs how the factors \emph{inside} a candidate decomposition combine --- whether they reinforce one another (constructive intra-cluster interaction, raising the cluster score) or attenuate one another (destructive, lowering it).

\textbf{Inter-cluster interaction} ($\kappa^{**}$). When two candidate clusters $C_1$ and $C_2$ both achieve plausibility against the same explanandum $\Phi$, a second-order question arises: are these \emph{compatible} decompositions (they share factors, or they identify complementary aspects of the phenomenon), or are they \emph{competing} decompositions (they propose mutually exclusive factor structures)? We introduce
\[
\kappa^{**}(C_1, C_2) \in [-1, 1]
\]
to encode this. Positive values indicate that $C_1$ and $C_2$ may be retained simultaneously as partial views of a richer decomposition; negative values indicate that committing to one would foreclose the other. The neutral case $\kappa^{**} \approx 0$ corresponds to clusters that address different aspects of $\Phi$ without bearing on one another.

Computationally, $\kappa^{**}$ can be derived from the structural overlap of clusters: shared factors with aligned weights contribute positively; disjoint factor sets with similar projection profiles onto $\Phi$ suggest competition (each cluster claims to explain the same phenomenon through different decompositions). We leave the precise estimation procedure to specific instantiations --- in \S5 we develop a concrete realization for the cyber-adversarial setting.

\textbf{Threshold governance at both levels.} The commitment threshold $\tau$ governs derivation at either level. \emph{Cluster commitment} requires $\mathrm{sc}_S(C \mid \Phi) \geq \tau$ for the candidate cluster, together with the absence of competing plausible clusters $C'$ for which $\kappa^{**}(C, C') < 0$ and $\mathrm{sc}_S(C' \mid \Phi)$ is comparable. \emph{Decomposition commitment} --- committing to a particular decomposition over a competing one --- requires that $\kappa^{**}(C, C') < 0$ has stabilized and that $C$'s lead over $C'$ exceeds a separation margin calibrated to $\tau$. We make this explicit:

\begin{defn}[Commit-worthy decomposition]
\label{def:commit-worthy}
A candidate cluster $C$ is \emph{commit-worthy} relative to $\Phi$ and a set of competing clusters $\{C'_1, \ldots, C'_m\}$ if:
\begin{enumerate}
  \item $\mathrm{sc}_S(C \mid \Phi) \geq \tau$, and
  \item for every $C'_j$ with $\kappa^{**}(C, C'_j) < 0$ and $\mathrm{sc}_S(C'_j \mid \Phi) \geq \epsilon$,
  \[
  \mathrm{sc}_S(C \mid \Phi) - \mathrm{sc}_S(C'_j \mid \Phi) \;\geq\; \delta(\tau),
  \]
\end{enumerate}
where $\delta(\tau)$ is a separation margin that grows with $\tau$: higher-stakes decisions require larger margins of separation before commitment is warranted.
\end{defn}

This two-level architecture is the paper's central formal contribution. It captures a structural feature of analytic abduction that synthetic abduction does not encounter: in synthetic mode, hypothesis composites are constructed upward toward a single integrated explanation, and competition among composites is resolved by the same scalar dynamics. In analytic mode, two competing decompositions of the same explanandum may each be internally coherent --- each may exhibit strong intra-cluster reinforcement, each may project well onto $\Phi$ --- yet committing to one rather than the other determines which interventions the system will license. The threshold structure must therefore operate at both levels, and the inter-cluster term $\kappa^{**}$ provides the formal mechanism for representing decomposition rivalry as a first-class inferential object.

\subsection{Risk Asymmetry and the Cost of Misdecomposition}

The synthetic mode inherits its risk argument from \cite{pareschi2025qa}: premature commitment to a single composite hypothesis can foreclose the discovery of hybrid explanations that constructive interaction would otherwise reveal. The analytic mode inherits an analogous but structurally distinct argument.

In domains where action is licensed by the \emph{decomposition} --- public health response calibrated to the inferred transmission mechanism, financial intervention targeting the inferred risk channel, defensive countermeasures calibrated to the inferred adversary strategy --- acting on an incorrect decomposition does not merely produce a false explanation. It directs resources toward the wrong targets, and in pathological cases it actively \emph{worsens} outcomes by intervening on factors that are not in fact operative while leaving the actual causal structure undisturbed. A response to airborne transmission is ineffective if the mechanism is waterborne; a liquidity intervention is ineffective if the underlying risk is counterparty contagion; a perimeter hardening is ineffective if the adversary's operative strategy is supply-chain compromise.

This is \emph{causal misattribution risk}, and its formal signature is precisely the configuration that the $\kappa$--$\tau$ apparatus is designed to detect: $\mathrm{sc}_S(C \mid \Phi) \geq \tau$ holds for some candidate $C$, yet a competing $C'$ with $\kappa^{**}(C, C') < 0$ also achieves comparable plausibility. Under such conditions, the standard inferential imperative to \emph{select} would force premature commitment. The two-level threshold structure, with its separation-margin condition $\delta(\tau)$, instead enjoins continued investigation: the system has identified a plausible decomposition, but the evidence has not yet distinguished it from a structurally incompatible competitor.

The threshold $\tau$ thus encodes a domain-specific \emph{risk calibration}: where the cost of acting on the wrong decomposition vastly exceeds the cost of continued investigation, $\tau$ is set high and $\delta(\tau)$ is large, deferring commitment until the evidence definitively disambiguates competing decompositions. Where decomposition errors are recoverable and the cost of delay dominates, $\tau$ and $\delta(\tau)$ are set low, enabling rapid commitment. The logic itself does not prescribe these values; what it provides is the formal apparatus within which such calibrations become representable and inspectable.

\subsection{Summary of the Formal Apparatus}

The analytic mode is structured by four elements:

\begin{itemize}
  \item \textbf{The causal cluster} $C = \langle w, \kappa_C \rangle$ as the structured output of decomposition, preserving factor identity, weight, and internal interaction structure.
  \item \textbf{The cluster score} $\mathrm{sc}_S(C \mid \Phi)$, computed by reversing the synthetic operator $\otimes$ against the explanandum $\Phi$.
  \item \textbf{The two-level interaction architecture}, with $\kappa^*$ governing intra-cluster dynamics and $\kappa^{**}$ governing inter-cluster compatibility and competition.
  \item \textbf{The $\tau$-governed separation condition} $\delta(\tau)$, calibrating commitment to the asymmetry between investigation cost and the cost of acting on a wrong decomposition.
\end{itemize}
Sections 4 and 5 demonstrate this apparatus in two domains. The crisis-decomposition demonstration (\S4) illustrates the framework's operation in a non-adversarial setting where the principal risk is intervention misdirection. The adversarial demonstration (\S5) introduces a distinctive feature absent from non-adversarial cases: an adversary actively \emph{crafts} the explanandum $\Phi$ to mislead the analyst's causal decomposition. Under such conditions, premature commitment is not merely costly --- it becomes itself an attack surface that a sophisticated adversary can exploit.

The structural features just introduced --- suspended decomposition, the two-level interaction architecture, separation margins --- collectively enable a distinctive mode of human-AI engagement that \S6 develops in detail: the legibility of suspended decomposition as a shared coordination object across multi-agent systems.

\section{Crisis Decomposition: The Epidemiological Case}

Crisis management is a paradigmatic domain for analytic abduction. The explanandum $\Phi$ --- an unfolding outbreak, an infrastructure failure cascade, a financial-system stress event --- presents itself as a complex pattern demanding decomposition. The factor library $\mathcal{F}$ comprises domain-specific causal mechanisms catalogued by expert knowledge. The risk profile is severely asymmetric: committing public-health resources to the wrong transmission model, hardening the wrong infrastructure node, or intervening on the wrong financial channel produces interventions that range from ineffective to actively harmful. And the temporal pressure is real: extended suspended decomposition is not free, because the phenomenon continues to evolve while commitment is deferred.

We develop the framework in the epidemiological setting. The example here is not adversarial --- the pathogen does not anticipate the epidemiologist's analytic model --- but it exhibits all the structural features that motivate the analytic mode: complex explanandum, heterogeneous factor library, multiple plausible decompositions, severe misdecomposition cost, and human-AI complementarity in cluster generation and threshold calibration. \S5 then introduces what the adversarial case adds that this case does not yet require.

\subsection{The Explanandum and the Factor Library}

Public-health authorities observe an outbreak of respiratory illness over an eight-week window, with three structured features: (i) initial geographic concentration in two adjacent metropolitan areas with high commuter overlap; (ii) age-stratified incidence showing elevated attack rates in working-age adults (25--55) compared to historical respiratory-virus baselines; (iii) secondary spread patterns suggesting both household clustering and workplace-mediated transmission, with weak evidence of school-mediated chains despite open in-person schooling during the window.

The explanandum $\Phi$ is the integrated pattern: not the case counts in isolation, not the age distribution in isolation, but the joint structure of geography, demographics, and secondary-spread patterns presented as a single phenomenon requiring decomposition. As discussed in \S3.1, $\Phi$ is constituted upstream of the decomposition layer --- surveillance systems flag the anomalous pattern; public-health epidemiologists recognize it as decomposition-worthy. The analytic-abduction framework takes $\Phi$ as given and asks: which latent transmission factors, in which interaction structure, account for it?

Formally, the explanandum is encoded as the structured tuple $\Phi = \langle O_\Phi, R_\Phi, \mu_\Phi \rangle$ introduced in \S3.2. The constituent observations $O_\Phi$ include individual case records, geographic incidence counts, age-stratified attack rates, and secondary-spread indicators (household and workplace clustering events). The relational structure $R_\Phi$ encodes temporal ordering (case onset sequences), spatial proximity (case geography), demographic co-classification (age and occupation strata), and inferred transmission-chain membership where available from contact tracing. The salience measure $\mu_\Phi$ weights observations by epidemiological prominence: confirmed cases and well-characterized transmission chains receive high salience; weakly attributable cases receive lower salience. The relational aggregator $\Psi$ for the epidemiological setting can be instantiated as a temporal-spatial kernel that weights observation pairs by relational proximity in $R_\Phi$, ensuring that factor projections respect the structured pattern of the outbreak rather than treating cases as independent samples. The framework's projection $\alpha_i(\Phi)$ for each transmission factor $f_i$ then captures not merely the prevalence of features compatible with $f_i$ across the case set, but the \emph{structured fit} between the factor's expected transmission pattern and the relational topology of the observed outbreak.

The factor library $\mathcal{F}$ for respiratory-pathogen transmission is well-developed in epidemiological practice. The relevant subset for this explanandum includes:

\begin{itemize}
  \item $f_1$: transport-hub transmission (high-throughput commuter infrastructure),
  \item $f_2$: workplace transmission (sustained indoor co-presence among working-age adults),
  \item $f_3$: school transmission (child-mediated household introduction),
  \item $f_4$: household transmission (sustained close-contact among co-residents),
  \item $f_5$: mass-event transmission (concentrated short-duration exposure),
  \item $f_6$: healthcare-setting amplification (nosocomial spread),
  \item $f_7$: airborne long-range transmission in shared ventilation,
  \item $f_8$: behavioral-adherence heterogeneity (subpopulations with differential adherence to mitigation).
\end{itemize}
The factor library here is curated by public-health expertise; the catalogue exists prior to any particular outbreak. The analytic-decomposition task is to identify, given $\Phi$, which combination of these factors with what internal interaction structure best accounts for the observed pattern.

\subsection{Candidate Decompositions}

Following the workflow developed in \S5 for the adversarial case (LLM-assisted candidate generation, expert $\kappa$-elicitation, $\kappa$--$\tau$ scoring), the system proposes three candidate clusters as decompositions of $\Phi$:

$C_1 = \{f_1, f_2, f_4\}$, a \textbf{commuter-workplace coupling} decomposition: transport-hub mixing seeds workplaces, which become the primary transmission setting; household clustering follows as secondary spread from infected working-age index cases. The intra-cluster $\kappa_{C_1}$ reflects strong reinforcement between $f_1$ and $f_2$ (commuter mixing feeds workplace exposure), moderate reinforcement between $f_2$ and $f_4$ (working-age cases bring infection home), and weak direct interaction between $f_1$ and $f_4$ (households are downstream of workplaces, not directly of transport).

$C_2 = \{f_2, f_4, f_7\}$, a \textbf{ventilation-mediated workplace} decomposition: poor indoor ventilation in workplace settings drives sustained transmission, with household clustering as secondary effect; geographic concentration is explained by industry-specific occupancy patterns rather than commuter geography. The intra-cluster $\kappa_{C_2}$ reflects strong reinforcement between $f_2$ and $f_7$ (the ventilation factor \emph{amplifies} the workplace factor) and between $f_2$ and $f_4$ as before.

$C_3 = \{f_2, f_4, f_8\}$, a \textbf{behavioral-heterogeneity workplace} decomposition: the operative mechanism is workplace transmission \emph{concentrated within subpopulations with lower mitigation adherence}; geographic clustering reflects occupational concentration of these subpopulations rather than transport infrastructure or ventilation. The intra-cluster $\kappa_{C_3}$ reflects strong reinforcement between $f_2$ and $f_8$ (heterogeneous adherence within workplace settings) and the standard $f_2$--$f_4$ coupling.

All three decompositions share $f_2$ (workplace) and $f_4$ (household) --- both are unambiguously supported by the observed pattern --- but they differ in \emph{which additional factor} is invoked to explain the geographic structure and the absence of strong school-mediated chains. This is the analytic-mode signature: the decompositions agree on what is \emph{present} and disagree on what is \emph{operative}.

\subsection{Cluster Scoring and Inter-Cluster Compatibility}
\label{sec:epi-scoring}

As in the adversarial case (\S\ref{sec:worked-example}), the quantities below are \emph{computed} from the stated inputs by the same pipeline; only the domain inputs differ. The implementation and the exact numeric trace are provided as supplementary material.

\textbf{Computational realisation.} The explanandum's constituents are rendered as the observation set $O_\Phi = \{o_1,\dots,o_5\}$: $o_1$, initial geographic concentration in two adjacent metropolitan areas with high commuter overlap; $o_2$, elevated attack rates among working-age adults relative to historical respiratory baselines; $o_3$, secondary spread showing household clustering; $o_4$, secondary spread showing workplace-mediated transmission among working-age adults; $o_5$, weak school-mediated transmission chains despite open in-person schooling. The relational structure $R_\Phi$ is the outbreak's transmission-dependency graph, the edge set $\{(o_1,o_4),(o_4,o_2),(o_4,o_3),(o_2,o_3),(o_3,o_5)\}$; the salience $\mu_\Phi$ weights well-characterised features above weakly attributable ones ($\mu(o_5)$ lowest). Each $f_i$ and $o_j$ is embedded, common-component-removed, and turned into a per-observation activation by the logistic link of \S\ref{sec:worked-example} ($\beta=4$); projection uses the $R_\Phi$-blind aggregator $\Psi_{\min}$ and the $R_\Phi$-aware aggregator $\Psi_{\mathrm{rel}}$ ($\gamma=0.5$, $\beta'=6$); cluster scores use Eq.~\eqref{eq:cluster-score} ($\eta=0.3$) with the elicited $\kappa_C$ matrices of \S4.2. Commitment uses $\tau=0.70$ --- calibrated for public-health interventions where misdirection is costly but not catastrophic, and lower than the $\tau=0.75$ of the attribution setting because public-health countermeasures can often be deployed in parallel --- with separation margin $\delta(\tau)=\rho\tau$ ($\rho=0.133$, so $\delta(0.70)=0.093$) and floor $\epsilon=0.05$.

\textbf{Projection.} The activation matrix $a_{ij}$ (Table~\ref{tab:epi-projection}) is legible: the transport factor $f_1$ peaks on the commuter-geography observation $o_1$, the workplace factor $f_2$ on the workplace observation $o_4$, the school factor $f_3$ on the weak-school observation $o_5$, and the household factor $f_4$ on the household observation $o_3$. The ventilation factor $f_7$ projects weakly throughout --- the observations describe demographics and settings but nothing about the built environment --- which is what later places the ventilation decomposition lowest. As in the adversarial case, $\Psi_{\min}$ collapses the factors toward the mean while $\Psi_{\mathrm{rel}}$ recovers a discriminating projection (Table~\ref{tab:epi-psi}); all downstream quantities use $\alpha_i(\Phi)$ from $\Psi_{\mathrm{rel}}$.

\begin{table}[t]
\centering
\caption{Epidemiological case: per-observation activation $a_{ij}$ (logistic link, $\beta=4$, after common-component removal). Each factor's strongest observation is in bold.}
\label{tab:epi-projection}
\small
\begin{tabular}{lccccc}
\toprule
 & $o_1$ & $o_2$ & $o_3$ & $o_4$ & $o_5$ \\
\midrule
$f_1$ & \textbf{0.78} & 0.11 & 0.24 & 0.12 & 0.32 \\
$f_2$ & 0.19 & 0.71 & 0.46 & \textbf{0.88} & 0.64 \\
$f_3$ & 0.43 & 0.37 & 0.38 & 0.50 & \textbf{0.79} \\
$f_4$ & 0.64 & 0.56 & 0.72 & 0.28 & \textbf{0.74} \\
$f_5$ & \textbf{0.55} & 0.43 & 0.21 & 0.27 & 0.47 \\
$f_6$ & 0.16 & 0.38 & 0.60 & \textbf{0.60} & 0.32 \\
$f_7$ & \textbf{0.59} & 0.17 & 0.13 & 0.12 & 0.41 \\
$f_8$ & 0.31 & 0.44 & 0.40 & \textbf{0.66} & 0.28 \\
\bottomrule
\end{tabular}
\end{table}

\begin{table}[t]
\centering
\caption{Epidemiological case: projection $\alpha_i(\Phi)$ under the two aggregators.}
\label{tab:epi-psi}
\small
\begin{tabular}{lcc}
\toprule
factor & $\Psi_{\min}$ (R$_\Phi$-blind) & $\Psi_{\mathrm{rel}}$ (R$_\Phi$-aware) \\
\midrule
$f_1$ & 0.31 & 0.74 \\
$f_2$ & 0.58 & 0.98 \\
$f_3$ & 0.48 & 0.96 \\
$f_4$ & 0.57 & 0.99 \\
$f_5$ & 0.38 & 0.74 \\
$f_6$ & 0.42 & 0.97 \\
$f_7$ & 0.27 & 0.55 \\
$f_8$ & 0.43 & 0.97 \\
\bottomrule
\end{tabular}
\end{table}

\textbf{Cluster scores.} Applying Eq.~\eqref{eq:cluster-score} with weights $w_i \propto \alpha_i(\Phi)$ and the elicited $\kappa_C$ matrices:

\begin{center}
\small
\begin{tabular}{lccc}
\toprule
 & base$(C\mid\Phi)$ & coh$(C)$ & $\mathrm{sc}_S(C\mid\Phi)$ \\
\midrule
$C_1$ (commuter--workplace)      & 0.916 & $+0.372$ & $\mathbf{0.783}$ \\
$C_2$ (ventilation--workplace)   & 0.887 & $+0.399$ & $\mathbf{0.764}$ \\
$C_3$ (adherence--workplace)     & 0.977 & $+0.366$ & $\mathbf{0.834}$ \\
\bottomrule
\end{tabular}
\end{center}

All three clear $\tau=0.70$. The adherence-heterogeneity reading $C_3$ scores highest and the ventilation reading $C_2$ lowest, but the spread is narrow and --- as in the adversarial case --- the embedding channel does not by itself resolve which amplifier of workplace transmission is operative.

\textbf{Inter-cluster compatibility.} The computed structural prior is mildly positive for all pairs ($\kappa^{**}_{\mathrm{struct}}(C_1,C_2)=+0.24$, $\kappa^{**}_{\mathrm{struct}}(C_1,C_3)=+0.20$, $\kappa^{**}_{\mathrm{struct}}(C_2,C_3)=+0.22$): the three decompositions share the workplace--household core and overlap heavily. The operational rivalry --- that committing to one amplifier deprioritises interventions targeting the others --- is supplied by elicitation: $\kappa^{**}(C_1,C_2)=-0.2$ (weakly competing; workplace-targeted measures address both), $\kappa^{**}(C_1,C_3)=-0.4$ (commuter-infrastructure measures versus adherence campaigns are substantially different interventions), $\kappa^{**}(C_2,C_3)=-0.3$. The milder operational incompatibility of the public-health setting, relative to the attribution case, is carried by these smaller $|\kappa^{**}|$ magnitudes rather than by a smaller separation margin.

\textbf{Separation analysis and suspended decomposition.} With $\tau=0.70$ and $\delta(\tau)=0.093$, the leading reading $C_3$ leads its nearest operationally incompatible competitor $C_1$ by $0.834-0.783=0.051$, and $C_1$ leads $C_2$ by $0.019$; both gaps lie below $\delta(\tau)$. By Definition~\ref{def:commit-worthy} no decomposition is commit-worthy, and the framework returns:

\begin{quote}
\emph{Three plausible decompositions identified, distinguished primarily by the inferred amplifier of workplace transmission: adherence heterogeneity ($C_3$, $0.834$), commuter mixing ($C_1$, $0.783$), and ventilation ($C_2$, $0.764$). All clear $\tau$ and all share workplace and household factors as central, but separation among them is below $\delta(\tau)=0.093$ and they are operationally incompatible ($\kappa^{**}<0$). Commitment suspended. Disambiguating evidence required: occupational-sector stratification of incidence (distinguishes $C_3$); built-environment ventilation assessment of high-incidence workplaces (distinguishes $C_2$); origin--destination commuter-flow analysis correlated with case onset (distinguishes $C_1$). Provisional intervention guidance: measures consistent with all three decompositions (workplace ventilation improvement, sick-leave policy) can be deployed without commitment; measures specific to a single decomposition (commuter-infrastructure modification, targeted-adherence campaigns) await disambiguation.}
\end{quote}

\subsection{The Distinctive Output: Suspended Decomposition as Action Guidance}

Conventional analytic practice in public-health crisis response, faced with three competing candidate decompositions and insufficient evidence to choose among them, has two unsatisfactory options. The first is \emph{premature commitment}: select the highest-scoring decomposition and deploy interventions accordingly. This is the failure mode the framework is designed to prevent --- the operative mechanism may be $C_3$ while the response targets $C_1$, and the misdirection of resources is the harm. The second is \emph{paralysis}: defer all action until disambiguation, accepting continued spread as the cost of inferential caution.

Analytic abduction under $\kappa$--$\tau$ governance offers a third option that neither alternative supports: \emph{commitment-stratified intervention}. The framework's output identifies which interventions are warranted under \emph{every} plausible decomposition (and can therefore be deployed without commitment to any of them), and which interventions are specific to a single decomposition (and should be deferred until disambiguation). The suspended-decomposition state is not a non-decision; it is a \emph{structured} decision that selects interventions robust to the unresolved ambiguity while flagging the evidence that would license further commitment.

This is the distinctive contribution of analytic abduction to crisis decision-making: not a faster answer, but a more \emph{decision-relevant} characterization of the current inferential state. The state of partial knowledge becomes itself actionable, in a way that scalar-confidence outputs and probability distributions do not support.

\subsection{What the Adversarial Case Will Add}

The epidemiological setting exhibits all the features that motivate analytic abduction in its general form: complex explanandum, multiple plausible decompositions, severe misdecomposition cost, and the value of legible suspended decomposition for action guidance. But it does not yet exhibit one feature that becomes structurally central in adversarial settings: the explanandum itself is not strategically generated. The pathogen does not anticipate epidemiological reasoning; the transmission pattern is causally generated by underlying mechanisms that do not respond to being analyzed.

In adversarial settings, this generative innocence is lost. The next section develops the analytic mode in which the explanandum is itself the product of an adversary's strategic choice --- including the choice of what should be analytically observable. Under such conditions, premature commitment is not merely costly: it becomes an attack surface, and the legibility of suspended decomposition becomes both a defense against exploitation and a guide to action.

\section{Adversarial Strategy Decomposition}

The crisis-decomposition demonstration of \S4 illustrated analytic abduction in a setting where the principal risk is \emph{intervention misdirection}: acting on the wrong decomposition wastes resources and may worsen outcomes, but the phenomenon being decomposed does not itself respond to the analyst's reasoning. In adversarial settings, this assumption breaks. The phenomenon under decomposition is the observable behavior of an opponent whose reasoning anticipates the analyst's decomposition. Premature commitment is not merely costly --- it becomes an \emph{attack surface}, exploitable by an adversary sophisticated enough to model the analyst's causal attribution and craft observable behavior to mislead it.

This section develops the analytic mode of QA in adversarial reasoning, with cyber threat analysis as the concrete domain. The setting brings into sharp focus three structural features of the framework that the non-adversarial case underuses: the cost-asymmetry of misdecomposition is operationally severe; the factor library $\mathcal{F}$ is shared by analyst and adversary, since the adversary reasons over the same vocabulary of strategic principles; and a phenomenon we will call \emph{structural-versus-linguistic novelty} --- the dissociation between novelty in the surface vocabulary and novelty in the compositional structure --- emerges naturally as a $\kappa$-theoretic configuration with operational significance.

\subsection{Setting: Decomposing Observed Adversary Behavior}

A defender observes a sequence of indicators attributable to a single adversary campaign: tactics, techniques, and procedures (TTPs) drawn from a shared catalogue such as MITRE ATT\&CK \cite{mitre-attack}, together with their timing, targeting, and operational signatures. The integrated explanandum $\Phi$ is the campaign-as-observed: not the individual TTPs in isolation, but the structured pattern they jointly exhibit --- sequencing, escalation, evasion choices, target selection. The defender's task is to \emph{decompose} $\Phi$: to identify the latent strategic principles whose interaction accounts for the observed campaign, and to do so under the constraint that committing to a wrong decomposition will misdirect defensive countermeasures.

Formally, the explanandum is encoded as the structured tuple $\Phi = \langle O_\Phi, R_\Phi, \mu_\Phi \rangle$ introduced in \S3.2. The constituent observations $O_\Phi$ are the individual indicators observed across the campaign: TTPs drawn from MITRE ATT\&CK, host and network artifacts, targeting events, and operational-signature observations. The relational structure $R_\Phi$ encodes temporal sequencing (which TTPs precede or follow which others), operational dependency (which observations are technically required for which subsequent ones), and target-victim coupling (which observations are directed at which assets). The salience measure $\mu_\Phi$ weights observations by analytic prominence: high-confidence indicators tied to ground-truth telemetry receive high salience; speculative attributions or low-confidence indicators receive lower salience. The relational aggregator $\Psi$ for the cyber setting can be instantiated as a sequence-aware kernel that weights observation pairs by their position in the inferred operational chain --- observations that participate in the same operational dependency cluster contribute more strongly to factor projections than observations spread across uncoupled phases of the campaign. The framework's projection $\alpha_i(\Phi)$ for each strategic heuristic $f_i$ then captures whether the heuristic's expected operational signature is present in $\Phi$ \emph{with the relational topology characteristic of that heuristic's actual deployment}, not merely whether elements compatible with $f_i$ appear scattered across the campaign.

The factor library $\mathcal{F}$ for this setting is the catalogue of strategic heuristics developed in \cite{ghisellini2025entangled}, extended with operationally-grounded heuristics extracted from offensive practice in forthcoming work on entangled-heuristic offensive synthesis \cite{carapella}. Each heuristic $f_i \in \mathcal{F}$ is rendered as a conditional principle:

\begin{itemize}
  \item $f_1$: \emph{If perimeter defenses are mature, then prioritize indirect approach through supply chain or trusted relationships.}
  \item $f_2$: \emph{If attribution would invite retaliation, then craft observable behavior to mimic a different threat actor.}
  \item $f_3$: \emph{If detection latency is high, then favor time-delayed activation to decouple intrusion from impact.}
  \item $f_4$: \emph{If defender's analytic model is known or inferable, then exploit it through deceptive composition.}
\end{itemize}
The library is heterogeneous: some heuristics descend from classical military theory (Sun Tzu, Clausewitz, Liddell Hart) and are tactically agnostic; others are domain-specific operational principles extracted from offensive cybersecurity practice. The system embeds each $f_i$ as a vector in the shared semantic space (following \cite{ghisellini2025entangled,pareschi2025qa}; the choice of embedding model is a substitutable component, as discussed in \S2.1), enabling projection against $\Phi$.

The intra-cluster interaction $\kappa_C$ is initialized from semantic similarity following \cite{ghisellini2025entangled,pareschi2025qa}, with sign-and-magnitude adjustments elicited from offensive-security experts who understand how heuristics operationally reinforce or conflict in execution. Inter-cluster interaction $\kappa^{**}$ between candidate decompositions is computed as developed in \S3.5, with the cyber-specific addition that competing attributions to distinct threat actors typically yield $\kappa^{**} < 0$ --- committing to one attribution forecloses the other and licenses incompatible countermeasures.

\subsection{The Deception Problem}

In non-adversarial analytic settings, the explanandum $\Phi$ is causally generated by the underlying factor structure: an outbreak pattern is produced by transmission dynamics; a financial correlation pattern is produced by risk-channel coupling. The decomposition task is to invert this generative relation. The factors may be hard to discern, but they do not actively \emph{resist} being discerned.

In adversarial settings, this generative innocence is lost. The adversary's observable behavior $\Phi$ is \emph{itself} the product of strategic choice, including the strategic choice of \emph{what should be observable}. A sophisticated adversary anticipates the defender's analytic process and may deliberately craft $\Phi$ to misdirect decomposition. Three deception patterns are operationally significant:

\textbf{Mimicry.} The adversary crafts $\Phi$ to project strongly onto factors associated with a \emph{different} threat actor's known strategic profile. The defender's decomposition surfaces a cluster $C_{\mathrm{mimicked}}$ with high score $\mathrm{sc}_S(C_{\mathrm{mimicked}} \mid \Phi)$, and commitment to this attribution directs countermeasures appropriate to the impersonated actor rather than the actual one.

\textbf{Obfuscation.} The adversary crafts $\Phi$ to project weakly onto every coherent cluster, sustaining high inter-cluster $\kappa^{**}$ conflict that defers commitment indefinitely. The defender is forced into prolonged suspended decomposition not because the evidence is genuinely ambiguous but because the adversary has engineered the ambiguity.

\textbf{Misdirection by overcommitment.} The adversary identifies a plausible-but-incorrect decomposition that the defender is likely to commit to, and crafts $\Phi$ to \emph{strengthen} this decomposition's projection while the actual operative factors remain latent. The defender commits with confidence --- clears $\tau$ and the separation margin $\delta(\tau)$ --- and acts on a decomposition the adversary has constructed precisely to be acted upon.

The third pattern is the most consequential because it weaponizes the analyst's own commitment dynamics. The $\kappa$--$\tau$ apparatus does not eliminate this risk --- no inferential framework can fully neutralize an adversary modeling the framework itself --- but it provides structural defenses that probabilistic and eliminative frameworks lack. We develop these in \S5.3.

\subsection{Structural-versus-Linguistic Novelty}

A distinctive phenomenon emerges in the $\kappa$-theoretic analysis of adversarial decomposition, with implications beyond the cyber domain.

Consider an adversary campaign whose individual TTPs are entirely familiar --- every technique is documented in MITRE ATT\&CK, every operational principle has a parallel in the strategic heuristics literature. By the surface vocabulary alone, the campaign exhibits \emph{low linguistic novelty}: the defender's projection scores $\alpha_i = \cos(\mathbf{f}_i, \Phi)$ are individually high across multiple known factors. Conventional analytic systems --- which select the highest-projecting factors and synthesize them into an attribution --- produce a confident decomposition with strong activation across familiar elements.

Yet the \emph{combination} of these familiar factors may be structurally novel: the inferred intra-cluster interaction $\kappa_C$ exhibits a pattern that does not match any previously catalogued threat actor's compositional profile. The factors are not novel; their \emph{interaction structure} is. Formally, this corresponds to a configuration in which:

\begin{enumerate}
  \item $\mathrm{sc}_S(C \mid \Phi) \geq \tau$ is achievable for a candidate cluster $C$,
  \item the constituent factors all have high projection scores against $\Phi$,
  \item yet the $\kappa_C$ implied by the joint observation pattern is \emph{anomalous} with respect to the historical distribution of $\kappa_C$ patterns associated with known attribution candidates.
\end{enumerate}
The structured encoding of $\Phi$ introduced in \S3.2 supplies the formal anchor for this distinction. \emph{Linguistic novelty} is novelty at the level of $O_\Phi$: individual observations whose embeddings project weakly onto the factor library, $\cos(\mathbf{f}_i, \boldsymbol{o}_j)$ small for many $f_i$ and $o_j$. \emph{Structural novelty} is novelty at the level of $R_\Phi$: a relational topology among observations that does not match the relational patterns of any catalogued case, even when the observations themselves are individually familiar. The relational aggregator $\Psi$ propagates these two sources of novelty into the projection $\alpha_i(\Phi)$ in mathematically distinct ways. Linguistic novelty reduces $\Psi$'s individual inputs $\cos(\mathbf{f}_i, \boldsymbol{o}_j)$; structural novelty leaves these inputs intact but alters the way $\Psi$ combines them. A campaign exhibiting \emph{low linguistic but high structural novelty} therefore produces high individual projection values that nonetheless yield an anomalous joint projection profile under $\Psi$ --- and an inferred intra-cluster $\kappa_C$ pattern that, when constructed to reproduce this profile, does not match any known attribution baseline. This is the $\kappa$-theoretic signature: the input projections are familiar but their relational aggregation is not, and the cluster's interaction structure is therefore forced into an anomalous configuration.

We term this \emph{structural novelty without linguistic novelty}: an adversary using familiar tools in a novel compositional regime. The $\kappa$--$\tau$ apparatus detects this configuration as a specific signature --- high individual factor projection accompanied by anomalous $\kappa_C$ --- that warrants suspended decomposition even when scalar cluster scores would license commitment.

This is the analytic-mode counterpart of an intuition long present in expert threat analysis: \emph{the techniques look familiar, but something about how they fit together is wrong}. Senior analysts often resist confident attribution under precisely these conditions, and their resistance has historically been difficult to formalize. The $\kappa$-theoretic framing supplies the formalization: their intuition is reading the $\kappa_C$ pattern, not the individual projection scores, and the framework can make this reading explicit.

The operational consequence is that the system can issue an inspectable signal: \emph{``Plausible decomposition identified, but its internal interaction structure $\kappa_C$ is anomalous relative to known attribution patterns; suspending commitment pending investigation of compositional novelty.''} This signal is the structural-novelty-without-linguistic-novelty configuration made legible --- the system reports not just what it has not yet committed to, but \emph{why}, and what kind of further evidence would resolve the ambiguity.

A symmetric configuration also exists: \emph{linguistic novelty without structural novelty}. Here the individual factors have low projection scores (the surface TTPs do not match any familiar technique catalogue) but the inferred $\kappa_C$ matches a known compositional profile. This typically indicates a known adversary using new tools --- operationally less alarming than the inverse, because countermeasures calibrated to the strategic \emph{composition} remain effective even when the surface vocabulary shifts. The framework distinguishes these two cases, which conventional similarity-based attribution conflates.

\subsection{Worked Example}
\label{sec:worked-example}

We illustrate the apparatus on a stylized yet representative scenario drawn from public threat-intelligence reporting on multi-stage intrusions. In contrast to a purely schematic illustration, every quantity below is \emph{computed} from the stated inputs by the procedure of this subsection; the implementation and the exact numeric trace are provided as supplementary material,\footnote{An implementation reproducing every quantity, table, and figure in \S\S4--5 is available at \url{https://github.com/Aribertus/analytic-abduction}.} and the embedding layer is the only model-dependent component (it is substitutable, in the sense of \S2.1).

\textbf{The explanandum.} Over a six-month window, a defender observes a coordinated campaign with the following structured pattern: $o_1$, initial access through a compromised third-party software update; $o_2$, extended dormancy of approximately ninety days before any second-stage activity; $o_3$, credential harvesting targeting a narrow set of identity-management infrastructure components; $o_4$, lateral movement using legitimate administrative tooling, avoiding novel binaries; $o_5$, data staging in cloud storage services owned by the target organisation itself; $o_6$, exfiltration coinciding with routine high-volume backup operations to provide cover. The campaign is internally consistent, technically competent, and exhibits no obvious operational errors.

\textbf{The factor library.} The defender's library $\mathcal{F}$ includes heuristics from offensive practice. Among the factors with non-trivial projection against $\Phi$ are: $f_1$, supply-chain compromise as initial access vector; $f_2$, time-delayed activation to decouple intrusion from impact; $f_3$, living-off-the-land (legitimate tooling) to avoid detection; $f_4$, identity infrastructure as primary target; $f_5$, exfiltration cover through legitimate data flows; $f_6$, deception-of-attribution through technique mimicry; $f_7$, economically motivated cybercrime profile; $f_8$, nation-state intelligence collection profile.

\textbf{Candidate decompositions.} The LLM, prompted with $\Phi$ and given access to $\mathcal{F}$, proposes three candidate clusters:
$C_1 = \{f_1, f_2, f_3, f_4, f_5, f_8\}$, a \emph{patient nation-state intelligence operation};
$C_2 = \{f_1, f_3, f_4, f_5, f_7\}$, a \emph{sophisticated cybercrime operation} reading dormancy as detection-evasion rather than strategic patience;
$C_3 = \{f_1, f_2, f_4, f_6, f_8\}$, a \emph{nation-state actor mimicking cybercrime} ($f_6$ active) to deflect attribution. Each is supplied with an internal interaction matrix $\kappa_{C}$ recording the reinforcement structure its reading posits; the matrices are given in the supplement.

\textbf{Computational realisation.} The schematic operators of \S3 are instantiated as follows. Each $f_i$ and each $o_j$ is rendered as a short text and embedded (the reported run uses GloVe-6B-50d mean-pooled embeddings; any sentence encoder, e.g.\ Sentence-BERT \texttt{all-MiniLM-L6-v2}, plugs in through the same interface). Because mean-pooled embeddings of short, same-domain phrases are strongly anisotropic --- every pair cosine sits near $0.9$, so raw cosine cannot \emph{dis}confirm --- we remove the common component (the mean direction of the scenario vocabulary) and renormalise, following standard post-processing of distributional embeddings \cite{mu2018allbutthetop}. The per-observation activation is a logistic link $a_{ij} = \sigma\!\big(\beta\,\cos(\mathbf{f}_i,\boldsymbol{o}_j)\big)$ with temperature $\beta=4$. The relational structure $R_\Phi$ is the campaign's operational-dependency chain, the edge set $\{(o_1,o_2),(o_2,o_3),(o_3,o_4),(o_4,o_5),(o_5,o_6),(o_3,o_5)\}$. Cluster scores use Eq.~\eqref{eq:cluster-score} with coupling $\eta=0.3$; commitment uses $\tau=0.75$, separation margin $\delta(\tau)=\rho\tau$ with $\rho=0.133$ (so $\delta(0.75)=0.10$), and plausibility floor $\epsilon=0.05$.

\textbf{Projection and the role of $R_\Phi$.} Table~\ref{tab:projection} reports the activation matrix $a_{ij}$. The structure is legible and not hand-set: $f_1$ peaks on $o_1$ (supply chain $\leftrightarrow$ vendor update, $0.93$), $f_2$ on $o_2$ (dormancy, $0.95$), $f_4$ on $o_3$ (identity targeting, $0.82$), $f_5$ on $o_6$ (exfiltration cover, $0.87$), while the cybercrime factor $f_7$ projects weakly throughout. This is the per-observation evidence; the aggregator $\Psi$ turns it into a per-factor projection $\alpha_i(\Phi)$.

The two realisations of $\Psi$ behave very differently, and the difference is exactly the role of $R_\Phi$. The $R_\Phi$-blind aggregator
\[
\Psi_{\min}\!:\quad \alpha_i(\Phi) = \frac{\sum_j \mu_\Phi(o_j)\,a_{ij}}{\sum_j \mu_\Phi(o_j)}
\]
averages each factor over \emph{all} observations, including those it is silent on, and collapses every factor toward the mean (Table~\ref{tab:psi}, middle column): under $\Psi_{\min}$ the factors are nearly indistinguishable and no decomposition can be told from another. The $R_\Phi$-aware aggregator $\Psi_{\mathrm{rel}}$ instead (i) reinforces a factor's activation on $o_j$ when it also explains $o_j$'s $R_\Phi$-neighbours --- crediting coherent runs in the dependency chain rather than scattered isolated hits ---
\[
\tilde a_{ij} = \min\!\Big(1,\; a_{ij} + \gamma\!\!\sum_{(o_j,o_k)\in R_\Phi}\!\! \mu_\Phi(o_k)\,a_{ik}\Big),
\]
and (ii) pools over observations by salience-weighted soft-max attention, $\alpha_i(\Phi) = \sum_j s_{ij}\tilde a_{ij}$ with $s_{ij}\propto \mu_\Phi(o_j)\exp(\beta'\tilde a_{ij})$, so a strategic factor is scored on the stage it actually explains rather than diluted to the mean ($\gamma=0.5$, $\beta'=6$). The result (Table~\ref{tab:psi}, right column) sharpens the projections and is what makes the structural/linguistic novelty distinction of \S5.3 computable: $\Psi_{\min}$ discards $R_\Phi$ by construction, $\Psi_{\mathrm{rel}}$ exploits it. All downstream quantities use $\alpha_i(\Phi)$ from $\Psi_{\mathrm{rel}}$.

\begin{table}[t]
\centering
\caption{Per-observation activation $a_{ij}=\sigma(\beta\cos(\mathbf{f}_i,\boldsymbol{o}_j))$ after common-component removal ($\beta=4$). Each factor's strongest observation is shown in bold; the matrix exhibits the expected factor-to-stage correspondence and the uniformly weak cybercrime row $f_7$.}
\label{tab:projection}
\small
\begin{tabular}{lcccccc}
\toprule
 & $o_1$ & $o_2$ & $o_3$ & $o_4$ & $o_5$ & $o_6$ \\
\midrule
$f_1$ & \textbf{0.93} & 0.32 & 0.62 & 0.35 & 0.78 & 0.37 \\
$f_2$ & 0.27 & \textbf{0.95} & 0.10 & 0.34 & 0.32 & 0.73 \\
$f_3$ & 0.34 & 0.45 & 0.45 & \textbf{0.78} & 0.31 & 0.32 \\
$f_4$ & 0.54 & 0.20 & \textbf{0.82} & 0.36 & 0.79 & 0.20 \\
$f_5$ & 0.26 & 0.40 & 0.51 & 0.56 & 0.54 & \textbf{0.87} \\
$f_6$ & 0.27 & 0.46 & 0.36 & \textbf{0.70} & 0.24 & 0.24 \\
$f_7$ & 0.27 & 0.28 & \textbf{0.58} & 0.38 & 0.40 & 0.19 \\
$f_8$ & 0.40 & 0.28 & 0.43 & 0.27 & \textbf{0.44} & 0.37 \\
\bottomrule
\end{tabular}
\end{table}

\begin{table}[t]
\centering
\caption{Projection $\alpha_i(\Phi)$ under the two aggregators. $\Psi_{\min}$ (which ignores $R_\Phi$) collapses the factors toward the mean and cannot separate them; $\Psi_{\mathrm{rel}}$ (which exploits $R_\Phi$) recovers a discriminating projection.}
\label{tab:psi}
\small
\begin{tabular}{lcc}
\toprule
factor & $\Psi_{\min}$ (R$_\Phi$-blind) & $\Psi_{\mathrm{rel}}$ (R$_\Phi$-aware) \\
\midrule
$f_1$ & 0.58 & 0.99 \\
$f_2$ & 0.43 & 0.88 \\
$f_3$ & 0.44 & 0.97 \\
$f_4$ & 0.50 & 0.97 \\
$f_5$ & 0.51 & 0.98 \\
$f_6$ & 0.38 & 0.91 \\
$f_7$ & 0.36 & 0.93 \\
$f_8$ & 0.37 & 0.82 \\
\bottomrule
\end{tabular}
\end{table}

\textbf{Cluster scores.} Applying Eq.~\eqref{eq:cluster-score} with the within-cluster weights $w_i \propto \alpha_i(\Phi)$ and the elicited $\kappa_{C}$ matrices yields

\begin{center}
\small
\begin{tabular}{lccc}
\toprule
 & base$(C\mid\Phi)$ & coh$(C)$ & $\mathrm{sc}_S(C\mid\Phi)$ \\
\midrule
$C_1$ (patient nation-state)      & 0.938 & $+0.416$ & $\mathbf{0.812}$ \\
$C_2$ (sophisticated cybercrime)  & 0.967 & $+0.396$ & $\mathbf{0.832}$ \\
$C_3$ (nation-state mimicking crime) & 0.917 & $+0.442$ & $\mathbf{0.799}$ \\
\bottomrule
\end{tabular}
\end{center}

All three clear the threshold $\tau=0.75$. Note that the cybercrime reading $C_2$ in fact attains the \emph{highest} score: at the resolution of the embedding channel the three decompositions are nearly tied, and the channel does not by itself separate the operationally critical nation-state/cybercrime distinction. A system that committed to its top-scoring decomposition would therefore commit, wrongly, to $C_2$. This is precisely the failure the governance layer exists to prevent.

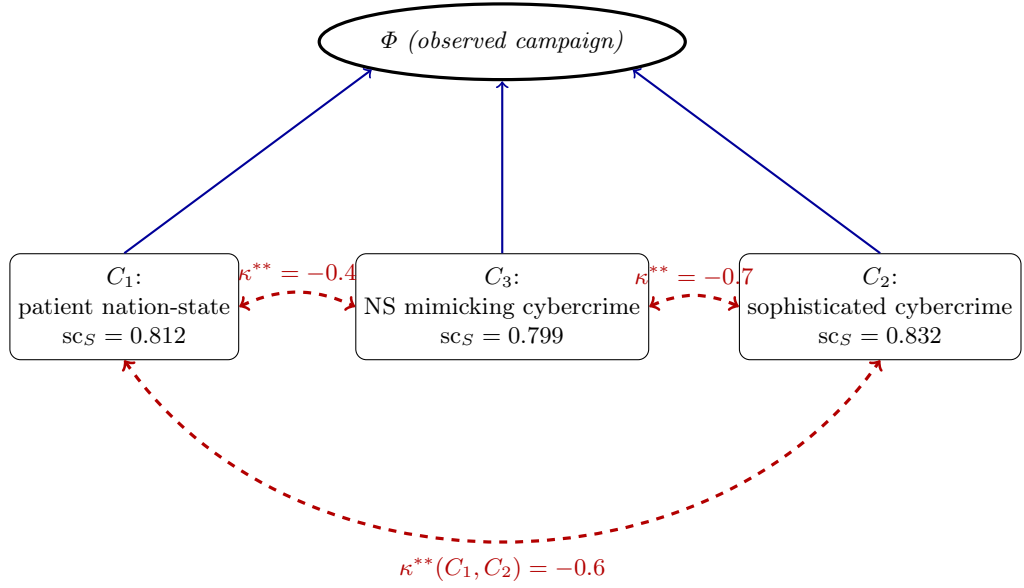
\begin{figure}[t]
\centering
\begin{tikzpicture}[
  cluster/.style={rectangle, draw, rounded corners, minimum width=3.0cm, minimum height=1.4cm, align=center, font=\small},
  phi/.style={ellipse, draw, very thick, minimum width=2.8cm, minimum height=1.0cm, font=\small\itshape},
  proj/.style={->, thick, blue!60!black},
  comp/.style={<->, very thick, red!70!black, dashed},
]
\node[phi] (phi) at (5, 3.5) {$\Phi$ (observed campaign)};
\node[cluster] (C1) at (0, 0) {$C_1$: \\ patient nation-state \\ \footnotesize $\mathrm{sc}_S = 0.812$};
\node[cluster] (C3) at (5, 0) {$C_3$: \\ NS mimicking cybercrime \\ \footnotesize $\mathrm{sc}_S = 0.799$};
\node[cluster] (C2) at (10, 0) {$C_2$: \\ sophisticated cybercrime \\ \footnotesize $\mathrm{sc}_S = 0.832$};
\draw[proj] (C1.north) -- (phi.south west);
\draw[proj] (C3.north) -- (phi.south);
\draw[proj] (C2.north) -- (phi.south east);
\draw[comp] (C1.east) to[bend left=25] node[above, font=\footnotesize] {$\kappa^{**} = -0.4$} (C3.west);
\draw[comp] (C3.east) to[bend left=25] node[above, font=\footnotesize] {$\kappa^{**} = -0.7$} (C2.west);
\draw[comp] (C1.south) to[bend right=55] node[below=2pt, font=\footnotesize] {$\kappa^{**}(C_1, C_2) = -0.6$} (C2.south);
\end{tikzpicture}
\caption{Computed worked example. Three candidate decompositions project against the observed campaign $\Phi$ with the cluster scores shown; all three clear $\tau=0.75$. Dashed red edges are the elicited inter-cluster compatibilities $\kappa^{**}$: all three pairs are operationally incompatible. The top-scoring decomposition is the cybercrime reading $C_2$, but its lead over the incompatible $C_1$ is only $0.020$, far below the separation margin $\delta(\tau)=0.10$; no decomposition is commit-worthy and the framework reports suspended decomposition.}
\label{fig:cyber-worked-example}
\end{figure}

\textbf{Inter-cluster compatibility.} The structural prior $\kappa^{**}_{\mathrm{struct}}(C_a,C_b)=\mathrm{agreement}-\mathrm{rivalry}$, with $\mathrm{agreement}=\sum_{f\in F_a\cap F_b}\min(w_a(f),w_b(f))$ and $\mathrm{rivalry}=(1-J)\,(1-|\mathrm{base}_a-\mathrm{base}_b|)$ for Jaccard overlap $J$, is computed as $\kappa^{**}_{\mathrm{struct}}(C_1,C_2)=+0.28$, $\kappa^{**}_{\mathrm{struct}}(C_1,C_3)=+0.23$, $\kappa^{**}_{\mathrm{struct}}(C_2,C_3)=-0.31$. The structural prior sees $C_1$ and $C_3$ as \emph{compatible} (both nation-state, high factor overlap) --- which is exactly why it is insufficient. The operationally decisive incompatibility is supplied by elicitation (\S5.5): nation-state and cybercrime attributions license disjoint countermeasures, and $C_3$ brands $C_2$'s very profile as fabricated. The elicited values used in commitment are $\kappa^{**}(C_1,C_2)=-0.6$, $\kappa^{**}(C_1,C_3)=-0.4$, $\kappa^{**}(C_2,C_3)=-0.7$ (Figure~\ref{fig:cyber-worked-example}).

\textbf{Separation analysis and suspended decomposition.} With $\tau=0.75$ and $\delta(\tau)=0.10$, the leading decomposition $C_2$ leads its nearest operationally incompatible competitor $C_1$ by $0.832-0.812=0.020$, and $C_1$ leads $C_3$ by $0.013$; both gaps lie far below $\delta(\tau)$. By Definition~\ref{def:commit-worthy} no candidate is commit-worthy, and the framework returns:

\begin{quote}
\emph{Plausible decompositions identified: $C_2$ (sophisticated cybercrime, $0.832$), $C_1$ (patient nation-state, $0.812$), $C_3$ (nation-state mimicking cybercrime, $0.799$). All clear $\tau$, but the top three are separated by less than $\delta(\tau)=0.10$ and are pairwise operationally incompatible ($\kappa^{**}<0$). Commitment suspended. Disambiguation requires evidence bearing on (i) monetisation versus collection --- access maintenance without exploitation would favour the nation-state readings over $C_2$; and (ii) independent corroboration of the mimicry hypothesis $f_6$ --- communications-pattern or targeting evidence beyond identity systems would separate $C_3$ from $C_1$.}
\end{quote}

This output is \emph{the} deliverable of the framework in this case. It is not an answer but a structured account of an ongoing inquiry, naming the live decompositions, the reason no commitment is yet warranted, and the specific evidence that would resolve it.

\textbf{Provenance of the quantities.} The example mixes four kinds of input, and the framework keeps them distinct (Table~\ref{tab:provenance}); this separation is itself part of what makes the output legible.

\begin{table}[t]
\centering
\caption{Provenance of the quantities in the worked example. The framework's legibility depends on keeping the computed, elicited, assumed, and institutional inputs distinct and separately inspectable.}
\label{tab:provenance}
\small
\begin{tabular}{ll}
\toprule
quantity & provenance \\
\midrule
embeddings, $a_{ij}$, $\alpha_i(\Phi)$, base, coh, $\mathrm{sc}_S$ & computed (embedding model + Eqs.) \\
$\kappa^{**}_{\mathrm{struct}}$ & computed (structural-overlap formula) \\
factor library $\mathcal{F}$, candidate clusters, $\kappa_C$ & elicited (LLM proposal + analyst) \\
salience $\mu_\Phi$, relational structure $R_\Phi$ & elicited (analyst / upstream systems) \\
$\kappa^{**}$ operational incompatibility & elicited (analyst calibration) \\
$\beta,\gamma,\beta',\eta$ & assumed (stated modelling constants) \\
$\tau$, $\delta(\tau)$, $\epsilon$ & institutional (stakes calibration) \\
\bottomrule
\end{tabular}
\end{table}

\textbf{Linguistic versus structural novelty in the computed example.} The run instantiates the dissociation of \S5.3. Under $\Psi_{\mathrm{rel}}$ every factor projects strongly ($\alpha_i \geq 0.82$): the campaign exhibits \emph{low linguistic novelty}, being assembled from individually familiar factors. What the framework cannot resolve is which \emph{interaction structure} generated it --- whether $f_6$ (mimicry) is operative, i.e.\ whether $C_3$ rather than $C_1$ holds. That distinction lives in $\kappa_C$, not in the projections, and separating it would require comparing the cluster's interaction structure against a corpus of catalogued actor $\kappa_C$-baselines, which the present framework does not assume. The honest output is therefore suspension: the framework reports that the decompositions are individually plausible, mutually incompatible, and not separable by the evidence channel available to it --- and names the structural evidence that would separate them. Full $R_\Phi$-sensitive structural-novelty detection is left to future work (\S8).

\subsection{Human-AI Coordination Within the Demonstration}

The worked example exhibits the loci of human-AI interaction identified in \S3 and previewed for \S6:

\textbf{Cluster generation.} The LLM proposes the candidate clusters $C_1, C_2, C_3$, drawing on its training over strategic and operational literature to articulate decompositions in natural language that a human analyst can immediately understand and critique. Critically, the LLM is not selecting the \emph{correct} decomposition; it is generating a candidate set whose adequacy is then evaluated by the $\kappa$--$\tau$ machinery. Because the apparatus scores only the candidates it is given, the framework's adequacy depends on the \emph{recall} of this generation stage: a decomposition absent from the candidate set cannot be scored, and a suspended verdict certifies only that no \emph{supplied} candidate separated, not that the operative decomposition was necessarily among them. In practice the candidate set is enlarged until coverage stabilises, spurious factors are filtered by their low projection $\alpha_i(\Phi)$ against the explanandum, and the human analyst can introduce candidates the LLM omits; treating generation recall as a measurable quantity, and bounding the risk of a missing decomposition, is left to future work.

\textbf{$\kappa^{**}$ elicitation.} The inter-cluster compatibility values $\kappa^{**}(C_i, C_j)$ require expert judgment about operational incompatibility between attribution hypotheses. An LLM can propose initial estimates from semantic structure (nation-state and cybercrime profiles project onto different operational vocabularies and yield $\kappa^{**} < 0$ by default), but the \emph{magnitude} of incompatibility and its calibration to countermeasure-divergence is supplied by human analysts familiar with the response infrastructure.

\textbf{$\tau$-calibration.} The threshold $\tau = 0.75$ and separation margin $\delta(\tau) = 0.10$ are not technical parameters. They encode a judgment about the cost of acting on a wrong attribution --- a judgment that belongs to the institutional decision-making layer (incident-response leadership, threat-intelligence governance), not to the technical layer. The framework makes the threshold inspectable: an analyst can see \emph{what $\tau$ is}, ask \emph{why it is set there}, and reason about what would justify lowering or raising it.

\textbf{Suspended-decomposition inspection.} The framework's output is itself a coordination object. The analyst does not receive ``the answer is $C_1$''; the analyst receives a structured account of the current inferential state --- which decompositions are plausible, why no commitment is yet warranted, and what specific evidence would change the situation. This makes the system's \emph{non-commitment} legible in a way that conventional analytic systems do not support. The analyst can act on this output by directing further investigation toward the disambiguating evidence the framework has identified.

\textbf{Commitment authorization.} When sufficient disambiguating evidence accumulates and the framework reports a commit-worthy decomposition, the analyst still decides whether to act on it. The framework reports that commitment is now \emph{warranted}; the analyst decides whether commitment is \emph{strategically appropriate} given factors outside the framework's scope (diplomatic considerations, ongoing intelligence operations, response capacity). The role-separation is clean: the framework governs \emph{inferential} commitment; the human governs \emph{practical} commitment.

The next section generalizes these observations: the legibility of suspended decomposition is not merely a useful feature of analytic abduction in a single domain --- it constitutes a distinctive contribution to the structure of human-AI coordination, with implications that extend beyond cyber adversarial reasoning to any multi-agent decision setting under sustained uncertainty.

\section{Legible Suspended Decomposition as a Human-AI Coordination Mechanism}

The demonstrations of \S4 and \S5 illustrated analytic abduction in two domains. In each case, the framework's output was not a decomposition but a \emph{structured account of an ongoing inquiry}: which decompositions are plausible, which competitions are unresolved, what evidence would resolve them, and which provisional actions are warranted under the unresolved ambiguity. This section generalizes the observation. The legibility of suspended decomposition is not merely a useful feature of the framework in particular domains. It constitutes a distinctive contribution to the structure of human-AI coordination, with implications that extend to any multi-agent decision setting under sustained uncertainty.

\subsection{Premature Convergence as a Multi-Agent Failure Mode}

Multi-agent decision systems integrating human decision-makers, large language models, and specialized AI components face a coordination failure that single-agent systems do not exhibit. Each agent in such a system has its own characteristic processing speed, confidence profile, and risk calibration. Large language models, in particular, are predisposed by training to produce confident, fluent outputs even under conditions where uncertainty is the more accurate response. Specialized AI components --- classifiers, anomaly detectors, attribution systems --- typically report scalar confidences that compress hypothesis interaction into a single number. Human participants reason on slower timescales but bring contextual judgment and institutional knowledge that the AI components lack.

In the absence of structural mechanisms to govern collective commitment, whichever agent first reaches a confident conclusion drives the collective toward that conclusion. The LLM proposes an attribution; the specialized AI returns a high-confidence classification; the analyst, presented with two converging confident signals, treats the matter as settled. The collective commits not because the inferential state warrants commitment but because the \emph{fastest confident agent} has effectively voted for closure. We term this \emph{premature convergence}, and it is structurally distinct from the premature-commitment failure of single-agent reasoning: in the multi-agent case, the failure is driven not by an individual agent's overconfidence but by the absence of coordination mechanisms that distinguish \emph{individual confidence} from \emph{collective warrant for commitment}.

Premature convergence is particularly costly in the analytic-decomposition setting. A confident attribution by any single agent --- the LLM, a threat-intelligence classifier, an institutional expert --- can drive collective commitment to a decomposition that the $\kappa$--$\tau$ analysis would have flagged as suspended. The fastest confident voice closes the inquiry before the structurally legitimate non-commitment has been registered as such.

\subsection{Legibility as Structural Resistance}

The $\kappa$--$\tau$ apparatus offers a structural defense against premature convergence: the \emph{non-commitment state itself becomes a coordination object}. When the framework reports suspended decomposition with the structure exhibited in \S4.4 and \S5.4 --- plausible clusters identified, separation margins not met, disambiguating evidence specified --- this output is not merely informational. It is a shared representation of the current inferential state that every agent in the collective can read, contribute to, and reason about.

Three properties make this representation effective as a coordination mechanism.

\textbf{Explicit non-commitment.} The framework does not present a probability distribution or a confidence-ranked list; it presents the \emph{structural reasons} commitment has been deferred. An LLM that would otherwise produce a confident attribution can be conditioned on the suspended-decomposition state and produce, instead, a confident \emph{analysis of the ambiguity}. The agent's fluency is redirected from premature closure toward characterizing the inquiry's current shape.

\textbf{Inspectable parameters.} The threshold $\tau$, the separation margin $\delta(\tau)$, the intra-cluster $\kappa_C$ patterns, and the inter-cluster $\kappa^{**}$ values are all visible to every agent. A human analyst can ask \emph{why} commitment is suspended and receive a structural answer: not ``the model is 78\% confident'' but ``two competing decompositions have comparable cluster scores, their inter-cluster compatibility is $-0.4$, and the separation margin calibrated to this domain's risk profile is not met''. The reasons for non-commitment are themselves objects of inquiry.

\textbf{Actionable disambiguation.} The framework identifies, alongside the suspended decomposition, the \emph{evidence that would resolve it}. This converts non-commitment from a static state into a directed investigation: agents can prioritize their effort toward acquiring the disambiguating evidence the framework has flagged, rather than continuing to refine confident outputs over a fixed evidence base.

Each of these properties is structurally distinct from what scalar-confidence frameworks offer. A 78\%-confident classifier output gives no purchase for agents to contribute, dispute, or redirect; the only available response is to accept or reject the output. A suspended-decomposition output, by contrast, is a \emph{structured argument} in which agents at different points in the multi-agent system have well-defined roles: cluster proposal, $\kappa$-elicitation, $\tau$-calibration, evidence direction, and ultimate commitment authorization.

\subsection{Role Differentiation Across the Pipeline}

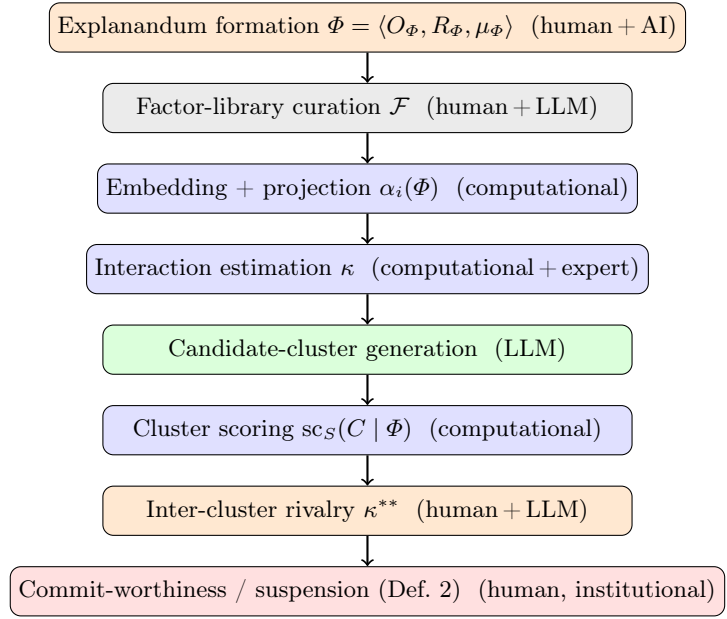
\begin{figure}[t]
\centering
\begin{tikzpicture}[
  stage/.style={draw, rounded corners, minimum width=7.0cm, minimum height=6.5mm, align=center, font=\small},
  comp/.style={stage, fill=blue!12},
  llm/.style={stage, fill=green!14},
  exp/.style={stage, fill=orange!18},
  inst/.style={stage, fill=red!12},
  collab/.style={stage, fill=black!8},
  arr/.style={->, thick},
  node distance=4mm
]
\node[exp] (phi) {Explanandum formation $\Phi=\langle O_\Phi,R_\Phi,\mu_\Phi\rangle$ ~{\footnotesize(human\,+\,AI)}};
\node[collab, below=of phi] (fac) {Factor-library curation $\mathcal{F}$ ~{\footnotesize(human\,+\,LLM)}};
\node[comp, below=of fac] (proj) {Embedding + projection $\alpha_i(\Phi)$ ~{\footnotesize(computational)}};
\node[comp, below=of proj] (kap) {Interaction estimation $\kappa$ ~{\footnotesize(computational\,+\,expert)}};
\node[llm, below=of kap] (gen) {Candidate-cluster generation ~{\footnotesize(LLM)}};
\node[comp, below=of gen] (sc) {Cluster scoring $\mathrm{sc}_S(C\mid\Phi)$ ~{\footnotesize(computational)}};
\node[exp, below=of sc] (riv) {Inter-cluster rivalry $\kappa^{**}$ ~{\footnotesize(human\,+\,LLM)}};
\node[inst, below=of riv] (com) {Commit-worthiness / suspension (Def.~\ref{def:commit-worthy}) ~{\footnotesize(human, institutional)}};
\foreach \a/\b in {phi/fac,fac/proj,proj/kap,kap/gen,gen/sc,sc/riv,riv/com}
  \draw[arr] (\a) -- (\b);
\end{tikzpicture}
\caption{The analytic-abduction pipeline, annotated by primary contributor. Computational stages (blue) compute projections and scores; the LLM (green) proposes candidate decompositions; human experts (orange) calibrate interaction structure and assess rivalry; institutional judgement (red) sets the commitment threshold and authorises commitment; explanandum and factor-library formation are collaborative (grey). The $\kappa$--$\tau$ apparatus evaluates the candidates the generation stage supplies; it does not itself generate them.}
\label{fig:pipeline}
\end{figure}

The demonstrations of \S4 and \S5 distributed human-AI contributions across the analytic-abduction pipeline. We make the distribution explicit here --- as a flow in Figure~\ref{fig:pipeline}, and stage by stage in Table~\ref{tab:role-differentiation}.

\begin{table}[t]
\centering
\caption{Role differentiation across the analytic-abduction pipeline.}
\label{tab:role-differentiation}
\begin{tabular}{p{3.4cm}p{3.4cm}p{5.0cm}}
\hline
\textbf{Locus} & \textbf{Primary contributor} & \textbf{Role} \\
\hline
Explanandum formation ($\Phi$) & Human + specialized AI & Recognize pattern complexity warranting decomposition \\
Factor library curation ($\mathcal{F}$) & Human + LLM & Mine and validate the catalogue of latent factors \\
Candidate cluster generation & LLM & Propose plausible decompositions in natural language \\
Intra-cluster $\kappa_C$ elicitation & Human expert & Calibrate operational interaction structure \\
Inter-cluster $\kappa^{**}$ elicitation & Human + LLM & Estimate decomposition compatibility \\
$\tau$ and $\delta(\tau)$ calibration & Human (institutional) & Encode risk profile of the domain \\
Suspended-decomposition inspection & Human + all AI agents & Read and contribute to the current inferential state \\
Disambiguation direction & All agents & Prioritize evidence acquisition \\
Commitment authorization & Human & Act on commit-worthy decomposition \\
\hline
\end{tabular}
\end{table}

This role distribution reflects two principles. First, \emph{AI contributes where its capabilities are distinctive}: LLMs excel at proposing candidate decompositions in natural language drawing on broad strategic and operational knowledge; specialized AI excels at detecting patterns and computing projections at scale. Second, \emph{humans retain authority where judgment is normatively required}: the calibration of $\tau$ encodes a risk judgment that belongs to institutional decision-making; commitment authorization is an act with consequences that humans must own.

Critically, the framework does not segregate ``AI tasks'' from ``human tasks''. The middle rows of the table --- $\kappa^{**}$ elicitation, suspended-decomposition inspection, disambiguation direction --- are explicitly \emph{collaborative}. The legibility of the framework's intermediate state is what makes this collaboration possible: agents at different points in the pipeline can read each other's contributions to the shared inferential state and respond to them.

\subsection{What Legibility Cannot Do}

The structural defenses described above are real but incomplete. Three limits warrant explicit acknowledgment.

The framework provides \emph{legibility}, not \emph{understanding}. The $\kappa$--$\tau$ apparatus exposes the structural features of the current inferential state in a form that domain experts can read. It does not guarantee that any particular human participant has the expertise to interpret what is exposed. A senior threat analyst can read a suspended-decomposition output and act on it; a less experienced participant may need supporting interface and explanation infrastructure. The framework enables expert engagement; it does not substitute for expertise.

The framework provides \emph{structural} resistance to premature convergence, not \emph{behavioral} resistance. An institutional culture that pressures rapid closure can override the framework's signals: a manager may demand a commitment that the framework reports as not commit-worthy, and the framework cannot prevent this. What it can do is make the override \emph{visible}: the decision to commit despite suspended decomposition becomes traceable, attributable, and accountable. Whether organizations make use of this traceability is outside the framework's scope.

The framework provides \emph{coordination}, not \emph{coordination guarantees}. Multiple agents reading the same suspended-decomposition output may still reach incompatible conclusions about what to do next. The framework structures the coordination space; it does not collapse it to a single answer. This is by design --- collapsing to a single answer would itself be a form of premature convergence --- but it means that multi-agent disagreement remains possible and must be managed through other mechanisms.

\subsection{Toward Architectural Realization}

The argument of this section has been principled rather than architectural: we have characterized what \emph{kind} of human-AI coordination the legibility of suspended decomposition enables, without specifying the system architecture within which it would be realized. Architectural realization requires a substrate in which agents of heterogeneous kind --- human decision-makers, LLMs, specialized AI components --- can read and contribute to a shared inferential state with well-defined interaction protocols. Multi-agent architectures developed for governed agentic systems \cite{borghoff2025frontiers,borghoff2025discover} provide candidate substrates, with their token-based coordination mechanisms naturally accommodating the structured outputs of analytic abduction. The development of an architectural realization --- translating the $\kappa$--$\tau$ apparatus into agent-level protocols, message types, and coordination primitives within such a substrate --- falls outside the scope of this paper and is the subject of forthcoming joint work.

What this paper has established is the \emph{target} for such an architecture: a multi-agent system in which suspended decomposition is a first-class shared object, in which the legibility of non-commitment is preserved across agent boundaries, and in which the structural resistance to premature convergence operates at the collective level rather than depending on the discipline of any single agent.

\section{Related Work}

The contribution of this paper is positioned at the intersection of several research traditions. We situate it briefly within each.

\textbf{Abductive reasoning in AI.} Classical abductive frameworks --- abductive logic programming~\cite{kakas1992abductive}, statistical abduction, set-covering theory~\cite{reggia1983setcovering}, and explanation-based reasoning --- can in principle return several competing explanations, and probabilistic approaches maintain distributions over hypotheses; the contrast we draw is therefore not that these frameworks necessarily select a single winner, but that none treats \emph{interaction among competing decompositions} and \emph{governed suspension of commitment} as first-class inferential objects. The most directly comparable methodology is the Analysis of Competing Hypotheses (ACH)~\cite{heuer1999psychology}, the established intelligence-analysis discipline for holding rival hypotheses open and scoring them against evidence; analytic abduction can be read as giving ACH a compositional substrate --- the interaction operator $\kappa$ --- and an explicit commitment calculus --- the threshold $\tau$ and the separation condition of Definition~\ref{def:commit-worthy} --- where ACH leaves both to unaided analyst judgement. The synthetic mode of \cite{pareschi2025qa} departs from the eliminative orientation by sustaining multiple hypotheses jointly and allowing them to reinforce one another, rather than eliminating all but one. The present paper extends this departure to the analytic direction: the structured object that survives suspended derivation is not a single hypothesis but a \emph{causal cluster} recording the decomposition's internal structure. Explanatory coherence frameworks in cognitive science~\cite{thagard2000coherence} share the present framework's emphasis on interaction structure among hypotheses, but operate on a network of explanatory relations rather than on an explicit compositional logic with governance thresholds.

\textbf{Causal reasoning under uncertainty.} Structural causal models~\cite{pearl2009causality} provide the dominant formalism for causal inference, with directed acyclic graphs, interventionist semantics, and the do-calculus supplying a powerful apparatus for reasoning about interventions on a known causal structure. The present framework addresses a different inferential question: not ``given this causal structure, what is the effect of intervention?'' but ``given this complex observed pattern, what causal structure accounts for it?''. Pearl's framework operates \emph{given} a causal graph; analytic abduction operates \emph{toward} one. In domains characterized by Knightian uncertainty~\cite{knight1921risk} --- where the causal structure itself is contested and probabilistic priors over candidate structures are unreliable --- the analytic mode provides a complementary inferential regime. The two approaches are most naturally combined sequentially: analytic abduction generates and disambiguates candidate causal structures; once a structure is commit-worthy, Pearl's framework licenses interventions over it.

\textbf{Adversarial reasoning and threat intelligence.} Cyber threat analysis has developed sophisticated taxonomies of adversary behavior, most notably the MITRE ATT\&CK framework~\cite{mitre-attack}, and a body of work on attribution that combines technical indicators with strategic and geopolitical analysis. Existing attribution practice typically operates by similarity matching against known actor profiles, supplemented by analyst judgment about strategic intent. Game-theoretic treatments of attribution additionally model the strategic interaction between an attacker who may craft deceptive indicators and a defender who anticipates such deception~\cite{attributiongames}; analytic abduction is complementary, supplying the inferential object --- the suspended decomposition --- over which such reasoning about evidence acquisition can proceed. The analytic-mode framework developed in \S5 contributes a structural extension: the inferred \emph{interaction pattern} among observed elements --- formalized as the intra-cluster $\kappa_C$ --- becomes itself an attribution signal, distinguishing structurally novel campaigns from linguistically novel ones in ways that vocabulary-level similarity matching cannot. Recent work on entangled-heuristic offensive synthesis~\cite{carapella} applies the synthetic mode of QA to the generation of operational cyber strategies; the present paper's analytic mode is the dual contribution --- decomposing observed offensive behavior into the latent strategic principles whose interaction generated it.

\textbf{Multi-agent reasoning and human-AI coordination.} Research on multi-agent decision systems has explored a range of coordination mechanisms, from auction-based and market-based approaches to consensus protocols and argumentation frameworks. Token-based coordination architectures developed for human-LLM-AI integration \cite{borghoff2025frontiers,borghoff2025discover} provide substrates within which heterogeneous agents can interact under structured protocols. The argument of \S6 is that the legibility of suspended decomposition contributes a distinctive coordination object to such architectures: a shared inferential state that is neither a commitment nor a confidence distribution but a structured representation of an ongoing inquiry. Within human-AI decision support specifically, the alternatives to a single committed answer are not limited to scalar confidence: calibrated uncertainty communication, ensemble disagreement, and explainable-AI interfaces for analysts all convey more than a point estimate~\cite{guidotti2018xai}. Suspended decomposition differs from these in kind rather than degree --- it reports not how uncertain a conclusion is, but which structured alternatives remain live, why commitment is withheld, and what would resolve it. The separation condition that withholds commitment in the presence of active, incompatible rivals has a suggestive parallel in conflict-monitoring accounts of human reasoning~\cite{deneys2012conflict}, where competing intuitions of comparable strength signal that deliberation should continue. This contribution is principled rather than architectural; the architectural realization within token-based coordination substrates is the subject of forthcoming joint work.

\textbf{Belief functions, possibility theory, and graded plausibility.} Frameworks for reasoning under uncertainty that depart from strict probabilistic norms --- Dempster-Shafer theory~\cite{shafer1976evidence} and its successors, possibilistic logic~\cite{dubois1988possibility}, ranking-theoretic approaches --- share with the present framework a recognition that probability is not the only or always the appropriate model of uncertainty. The $\kappa$--$\tau$ apparatus differs from these in two respects: the interaction operator $\kappa$ introduces compositional context-sensitivity among hypotheses that these frameworks do not represent, and the normative threshold $\tau$ structurally decouples inferential derivation from commitment in a way that these frameworks treat as external to the inferential apparatus. The present framework is thus closer in spirit to recent work on dialectical proof procedures~\cite{dung1995argumentation}, where non-termination is a first-class phenomenon --- though where dialectical frameworks treat non-termination as an obstacle to be managed, analytic abduction treats suspended decomposition as an epistemically legitimate and operationally productive state.

\section{Conclusion and Future Directions}

This paper has developed analytic abduction within the $\kappa$--$\tau$ apparatus: the dual orientation in which a complex observed state is decomposed into the latent factors whose interaction accounts for the observed complexity. The central formal contribution is the causal cluster as the structured output of decomposition, together with the two-level interaction architecture --- intra-cluster $\kappa^*$ and inter-cluster $\kappa^{**}$ --- and the $\tau$-governed separation condition that distinguishes commit-worthy decompositions from those that remain suspended. The framework was demonstrated in epidemiological crisis decomposition and in cyber adversarial reasoning, with the latter introducing the phenomenon of structural-versus-linguistic novelty as a $\kappa$-theoretic configuration of operational significance.

The paper's contribution to human-AI reasoning is the argument of \S6: that the legibility of suspended decomposition constitutes a distinctive coordination object across multi-agent systems integrating human decision-makers, large language models, and specialized AI components. This legibility provides structural resistance to premature convergence --- the failure mode in which the fastest confident agent drives collective commitment regardless of whether the inferential state warrants it --- by making non-commitment as inspectable, contestable, and actionable as commitment.

\subsection*{Limitations}

Four limitations qualify the contribution. First, the framework presupposes that the factor library $\mathcal{F}$ contains the operative factors or compositions thereof; decomposition under \emph{incomplete} factor libraries, where the operative factor is genuinely novel and not synthesizable from existing $\mathcal{F}$, falls outside the present framework. Second, the legibility argument of \S6 establishes what the framework \emph{enables} but not what it \emph{guarantees}: institutional cultures that pressure rapid closure can override the framework's signals, and the framework's contribution is to make such overrides visible rather than to prevent them. Third, the demonstrations of \S\S4--5 are stylized: the scenarios, the factor libraries, and the elicited inputs (the interaction matrices $\kappa_C$ and $\kappa^{**}$ and the salience $\mu_\Phi$) are illustrative rather than drawn from operational data. What is not illustrative is the computation over them --- the projections, cluster scores, and suspension decisions are produced by the reproducible pipeline of \S\ref{sec:worked-example} rather than asserted --- so the demonstrations establish the framework's structural operation without empirically validating its calibration to specific domains.

Fourth, the projections inherit the limitations of the embedding model that computes them. Off-the-shelf sentence encoders such as Sentence-BERT, especially untuned, handle some semantic phenomena poorly --- negation and contradiction most notably, where lexically similar but logically opposed statements can receive high cosine similarity --- and they resolve fine distinctions only coarsely (in the cyber example, the channel does not separate espionage from monetisation). Two things bound the consequences. The embedding layer is a substitutable, tunable component (\S2.1): a contradiction-aware or domain-adapted encoder can replace the default without any change to the $\kappa$--$\tau$ apparatus, and the common-component removal used in \S\S4--5 restores the channel's capacity to disconfirm. More fundamentally, the governance layer is designed so that an imperfect projection channel does not drive commitment: where the channel cannot cleanly separate competing decompositions, the framework suspends rather than forcing a choice, as both worked examples show. Embedding error thus degrades the sharpness of the projections, not the soundness of the suspension --- though a systematically biased encoder could still mislead the analyst about \emph{which} disambiguating evidence to seek, and domain adaptation of the embedding layer remains the principal lever for improving performance in any given setting.

\subsection*{Future Directions}

The framework opens several lines of development that fall naturally to distinct venues.

\emph{Architectural realization.} The legibility of suspended decomposition characterized in \S6 requires a multi-agent substrate within which heterogeneous agents can read and contribute to a shared inferential state with well-defined interaction protocols. Forthcoming joint work develops this realization within token-based coordination architectures \cite{borghoff2025frontiers,borghoff2025discover}, translating the $\kappa$--$\tau$ apparatus into agent-level protocols, message types, and the institutional governance structures that calibrate $\tau$ to domain risk profiles.

\emph{Domain instantiation in offensive and defensive cybersecurity.} The cyber demonstration of \S5 introduces the analytic mode's central concepts but does not develop them at the operational depth that cybersecurity practice requires. Forthcoming work, building on the offensive-strategic-synthesis pipeline developed in recent applied research, will deepen the structural-versus-linguistic novelty analysis as a defensive attribution signal and as a normative target for offensive operations seeking to evade premature defender commitment. The two directions are complementary: the offensive perspective clarifies which compositional configurations are achievable; the defensive perspective clarifies which configurations warrant suspended decomposition.

\emph{Logical foundations.} The $\kappa$--$\tau$ apparatus developed in \S2.3 and used throughout the paper admits a fuller proof-theoretic and semantic development. A dedicated treatment of the apparatus as a logical system --- with explicit language, scoring semantics, inference principles, and meta-theoretic results --- is the subject of separate work in preparation.

\emph{Completeness of the factor library.} The presupposition that $\mathcal{F}$ contains the operative factors is most fragile in genuinely novel crises --- emerging diseases, unprecedented adversary capabilities, novel financial instruments. Extending the analytic-mode framework to handle factor-library extension during an investigation --- where a candidate decomposition's failure to reach commit-worthy status signals not merely insufficient evidence but the absence of a needed factor --- is a substantive direction that intersects with research on creative hypothesis generation and conceptual blending.

\emph{Causal-graph integration.} The complementarity with Pearl's framework noted in \S7 admits formal development: a sequential pipeline in which analytic abduction produces a commit-worthy causal cluster $C$, which then licenses construction of a structural causal model over the factors in $C$, which then licenses interventionist reasoning. The integration would unite the inferential strengths of both frameworks --- analytic abduction's discipline under deep uncertainty, structural causal models' precision under known structure --- while preserving the role separation between them.

The framework presented here is, in the spirit of \cite{pareschi2025qa}, a substrate rather than a closed system. Its value lies in what it makes representable: the structured, ongoing state of reasoning under uncertainty, made legible to the multi-agent collectives in which such reasoning increasingly occurs.



\subsection*{Disclosure of Interests and Tools}

The author declares no competing interests relevant to this work. Large language models were used as assistive tools for drafting, editing, and implementing the reproducible computational pipeline of \S\S4--5; all claims, formal definitions, and analyses were authored and verified by the author, who takes full responsibility for the content. This assistive use is methodologically distinct from the framework's own use of an LLM as a candidate-generation component, described in \S\S5--6.

\end{document}